\documentclass[twocolumn,a4paper]{article}
\usepackage[top=1.5cm,bottom=2cm,left=1.3cm,right=1.3cm]{geometry}

\def\BibTeX{{\rm B\kern-.05em{\sc i\kern-.025em b}\kern-.08em
    T\kern-.1667em\lower.7ex\hbox{E}\kern-.125emX}}

\usepackage[T1]{fontenc}
\usepackage[utf8]{inputenc}
\usepackage{graphicx}
\usepackage{amsmath}
\usepackage{mathtools}
\usepackage{amsfonts}
\usepackage{algorithmicx}
\usepackage{algpseudocode}
\usepackage{algorithm}
\usepackage{subcaption}
\usepackage{multirow}
\usepackage[export]{adjustbox}
\usepackage{color}
\usepackage[dvipsnames]{xcolor}
\usepackage{hyperref}
\usepackage{authblk}

\begin{document}

\title{Detection of Binary Square Fiducial Markers Using an Event Camera}

\date{}
\author[2]{Hamid Sarmadi}
\author[1,2]{Rafael~Mu\~noz-Salinas}
\author[3]{Miguel A. Olivares-Mendez}
\author[1,2]{Rafael Medina-Carnicer}

\affil[1]{Computing and Numerical Analysis Department, University of C\'ordoba, Spain}
\affil[2]{The Maimonides Biomedical Research Institute of C\'ordoba (IMIBIC), Spain}
\affil[3]{Space Robotics Research Group, Interdisciplinary Centre for Security, Reliability and Trust (SnT), Universit\'e du Luxembourg, Luxembourg}

\maketitle
\begin{abstract}

Event cameras are a new type of image sensors that output changes in light intensity (events) instead of absolute intensity values. They have a very high temporal resolution and a high dynamic range. In this paper, we propose a method to detect and decode binary square markers using an event camera. We detect the edges of the markers by detecting line segments in an image created from events in the current packet. The line segments are combined to form marker candidates. The bit value of marker cells is decoded using the events on their borders. To the best of our knowledge, no other approach exists for detecting square binary markers directly from an event camera using only the CPU unit in real-time. Experimental results show that the performance of our proposal is much superior to the one from the RGB ArUco marker detector. The proposed method can achieve the real-time performance on a single CPU thread.

\end{abstract}

\section{Introduction}

Event cameras or ``silicon retina'' cameras \cite{lichtsteiner_128_2006} are a new type of image sensors with a fundamentally different approach to sensing. The name silicon retina comes from the similarity of these cameras to the retina in the human eye in the way they sense images. Instead of capturing absolute values for each pixel in the image, they sense the change in brightness at each pixel. This change in brightness is then compared to a threshold and if it is greater, an event for that pixel is produced and communicated. Positive changes in brightness produce the so-called \emph{on} events and negative changes in brightness produce the so-called \emph{off} events. Because of the asynchronous nature of these cameras, they are able to produce events with a very high temporal resolution (e.g. 1 microsecond) \cite{gallego_event-based_2019}. Another advantage of these cameras is their very high dynamic range which makes them suitable in situations with a very low or very high amount of illumination or with a high contrast of brightness in the field of view. They also tend to consume less energy compared to normal image sensors since they do not need to send a value for every pixel at regular intervals \cite{amir_low_2017}.

All these advantages have resulted in a lot of interest in developing different computer vision algorithms for event cameras \cite{kim_real-time_2016,maqueda_event-based_2018,vasco_fast_2016,bardow_simultaneous_2016,gallego_event-based_2019}. Although the currently available event cameras are more expensive and have relatively lower resolution compared to conventional counterparts, it is expected that they will become more available with higher resolutions in the future with companies like Samsung \cite{son_41_2017} investing in their mass production.

On the other hand, binary square fiducial markers are a popular technology for pose estimation in augmented reality and other applications. There are many examples of them such as IGD \cite{xiang_zhang_visual_2002}, Matrix \cite{rekimoto_matrix_1998}, binARyID \cite{froehlich_lightweight_2007}, ARToolKitPlus \cite{wagner_artoolkitplus_2007}, and ARTag \cite{fiala_designing_2010}. One of the most recent and most widely used versions is the ArUco marker \cite{garrido-jurado_generation_2016} which has been employed in important applications such as medicine and robotics \cite{dhall_lidar-camera_2017,sarmadi_3d_2019,su_improved_2019,sarmadi_joint_2020,bacik_autonomous_2017,sani_automatic_2017}. The ArUco markers are also used in more fundamental computer vision problems such as camera calibration, environment mapping, and SLAM \cite{sarmadi_simultaneous_2019, munoz-salinas_mapping_2018, munoz-salinas_ucoslam_2020}. All binary square fiducial markers have a system of ID codes which helps to identify unique markers within each type. These codes are presented on a square cell grid with black and white colors representing zeros and ones. This visual representation is common among all binary square markers. 

In this paper, we propose a method to detect and decode square binary markers using an event camera. The fundamental concept which we employ is that when a pixel moves from a white to a black binary grid cell it produces \emph{off} events and when it moves from a black to white grid cell produces \emph{on} events. Since binary square fiducial markers also have a black border around their code grid it is also possible to find the edge of the marker that is in the same direction as the marker's movement. This is done by finding a line segment of \emph{off} events and correspondingly a line segment of \emph{on} events on the opposite edge. 

To the best of our knowledge, this is the first work that attempts to decode fiducial planar markers in event cameras using only the CPU unit in real-time. It is possible to convert the events from the camera to an intensity image employing one of the recent methods \cite{brandli_real-time_2014,munda_real-time_2018,bardow_simultaneous_2016} and then use the RGB marker detection algorithms. This has been done in \mbox{\cite{holesovsky_practical_2019}} employing the intensity image reconstruction algorithm introduced in \mbox{\cite{rebecq_events--video_2019}}. However, these reconstruction methods either have a considerable amount of inaccuracy in intensity image reconstruction or they need a powerful dedicated GPU to perform their computations. On the other hand, there is another CPU-based method \mbox{\cite{nagata_qr-code_2020}} for detecting and decoding markers in event cameras that works based on optimizing a generative model. However it cannot operate in real-time despite it using a much more powerful hardware (10-Core CPU and 64 GB of RAM) than us. Our approach can run on a single CPU thread in real time which makes it more accessible and also suitable for low-powered devices.

The rest of this paper is structured as follows. Section \ref{sec:related_work} presents the related works, then in Section \ref{sec::methodology} the proposed approach is introduced in detail. Section \ref{sec::experiments} describes the experimental results and discussion for validating our method and finally Section \ref{sec:conclusion} draws some conclusions and suggests future work.
\begin{figure}[t!]
    \centering
    \includegraphics[width=0.5\textwidth]{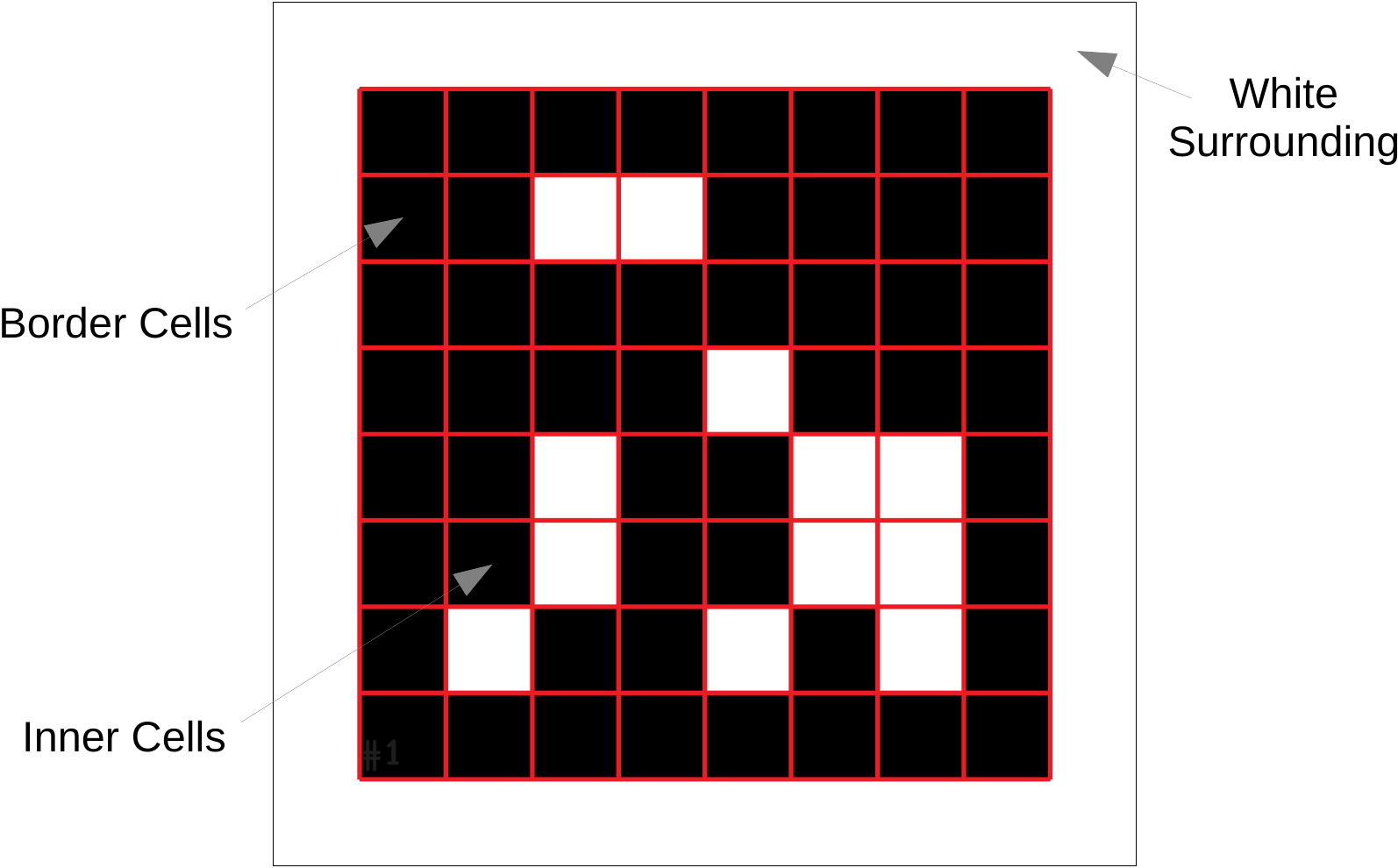}
    \caption{The structure of an ArUco marker. The marker is printed on a white background. The border cells of the marker are all black and the inner cells are used to keep the identification code using black and white colors. Please note that the red lines are for visualization of the cell grid and are not part of the marker.}
    \label{fig:aruco_structure}
\end{figure}

\section{Related Work}
\label{sec:related_work}

\subsection{Binary Square Marker Detection in Regular Cameras} 

Square binary markers are fiducial markers that can be easily detected in images and their pose can be efficiently determine with respect to the camera given the size of the square marker. One of their most important applications is augmented reality \cite{sarmadi_joint_2020,tabrizi_augmented_2015,sannikov_interactive_2015} however they are also used in other fundamental computer vision problems such as camera calibration \cite{atcheson_caltag_2010,fiala_self-identifying_2008}, 3D reconstruction \cite{degol_improved_2018,rumpler_automated_2014,sarmadi_3d_2019}, and SLAM \cite{munoz-salinas_spm-slam_2019,munoz-salinas_ucoslam_2020}. Additionally, they are applied to unmanned aerial vehicles \cite{sani_automatic_2017,bacik_autonomous_2017}\, medicine \cite{sarmadi_3d_2019,sarmadi_joint_2020}, and swarm robotics \cite{suzuki_shapebots_2019,fekrmandi_validation_2020} amongst other applications.

One of the first works to design a marker with binary squares for coding and a black edge for easy detection is Matrix \cite{rekimoto_matrix_1998}. To detect the marker, they first binarize the image and then select marker candidate areas by connected component analysis. They then fit a quadrangle to the area and using its four corners, the pixels are projected to a code frame. The code is then extracted by counting the white and black pixels in each cell. Finally, a CRC-error test is applied to the extracted code to check the validity of the marker. In another work by Fiala called ARTag \cite{fiala_artag_2005} marker candidates are detected by finding line segments in the extracted edge image. Where the lines segments form a quadrilateral a candidate is determined. For binarization, they use a threshold obtained from the edges of the marker. They also use CRC code correction to verify the code from the code cells. The set of possible markers is generated in a manner that minimizes the confusion between markers. In another method presented by Flohr and Fischer \cite{froehlich_lightweight_2007} a way of generating binary codes is introduced that does not confuse the markers with each other when the marker is rotated. In a more recent approach called ArUco by Garrido-jurado et. al. \cite{garrido-jurado_automatic_2014}, a probabilistic stochastic approach is employed to generate marker codes to maximize their distances from each other. For detection, the marker candidates are determined by contour extraction and then polygon approximation. The four-sided polygons are selected and a homography is calculated to transform them to the standard marker shape. For binarizing, an optimal bimodal image thresholding algorithm is employed. Finally, the extracted code from the cells is looked up in a dictionary to verify the marker ID.

\subsection{Event-based Cameras }
Event cameras are a new type of camera that senses changes in light intensity rather than their absolute value \cite{lichtsteiner_128_2006}. Since these sensors are fairly new, they are more expensive and have limited resolution compared to regular RGB cameras \cite{gallego_event-based_2019}. However, because of their different design, they are good at sensing with high temporal resolution and high dynamic range which makes them appropriate for outdoor and industrial applications. Researchers have already attempted to use these cameras for common computer vision problems which include object classification \cite{gao_end--end_2020}, image deblurring \cite{zhang_hybrid_2020}, face detection \cite{barua_direct_2016}, person tracking \cite{piatkowska_spatiotemporal_2012}, mosaicing \cite{kim_simultaneous_2014}, and 3D reconstruction \cite{kim_real-time_2016,zhou_semi-dense_2018}. 

One of the earliest examples of object detection and tracking by event cameras is the work of Litzenberger et. al. \cite{litzenberger_embedded_2006}. They cluster the events in the image spatially and update the clusters when new events arrive. This method is employed to track the cars on a highway. Another early work \cite{conradt_pencil_2009} used for balancing a pencil on its tip estimates the line representing the pencil in the image as a Gaussian in Hough space. They take advantage of a quadratic representation which is the log of the Gaussian. At the arrival of each event, a quadratic related to its position is added to the Hough space while the previous quadratic is slightly decayed. Later, Schraml et. al. \cite{schraml_dynamic_2010} presented a method that applied stereo matching to estimate depth using two event cameras. They introduced a tracking algorithm based on finding bounding boxes for each object where every bounding box contains events that are spatially connected. Piatkowska et. al. in \cite{piatkowska_spatiotemporal_2012} took advantage of a Gaussian mixture model to track multiple people in the scene using the maximum a posteriori probability estimation. After that, Reverter et. al. in \cite{valeiras_asynchronous_2015} tracked an object by assigning bivariate Gaussians to its different parts. The Gaussian trackers are restricted by spring-like links to track the object as a whole. They demonstrate their results for tracking of the human face. In another work by Glover et. al. \cite{glover_event-driven_2016} a moving ball is detected using a circular Hough transform that also integrates optical flow estimation to reduce false detections in a cluttered scene. Mitrokhin et al. \cite{mitrokhin_event-based_2018} minimize the background noise to track an object by a moving camera through iteratively optimizing a motion model employing the event count image and the event timestamp image.

Up to our knowledge, there are no attempts on using event cameras to detect and decode binary planar markers in real-time using only the CPU unit. There is a CPU based method presented in \mbox{\cite{nagata_qr-code_2020}} which decodes QR codes by optimizing a generative model. They first estimate and initialize the motion and affine transformation of the marker and then optimize these two parameters iteratively. At the end the QR code is optimized to be decoded. Nevertheless, they cannot achieve real-time detection and decoding in spite of employing a much more powerful hardware than us. On the other hand, one might use one of the methods for intensity image reconstruction from event cameras and then apply a normal marker detection algorithm as is done in \cite{holesovsky_practical_2019}. However, these methods have some drawback that we explain below. 
The first example is the work by Brandli et. al. \cite{brandli_real-time_2014}. They can get a high frame rate on CPU, however, their algorithm needs an intensity image of the first frame. Furthermore, there is a considerable amount of inaccuracy in their intensity reconstruction.
Another method which creates fairly acceptable reconstruction only taking advantage of events is presented by Munda et. al. \cite{munda_real-time_2018}. Their reconstruction is done on an event by event basis. They achieve real-time performance, however, they need to take advantage of GPU computing to achieve that.
Bradow et. al. present an algorithm in \cite{bardow_simultaneous_2016} which can reconstruct optical flow as well as intensity image. They use a sliding window-based algorithm and can achieve near real-time performance. However, they need a dedicated GPU to perform their optimization.

\begin{figure}[t!]
    \centering
    \includegraphics[width=0.35\textwidth]{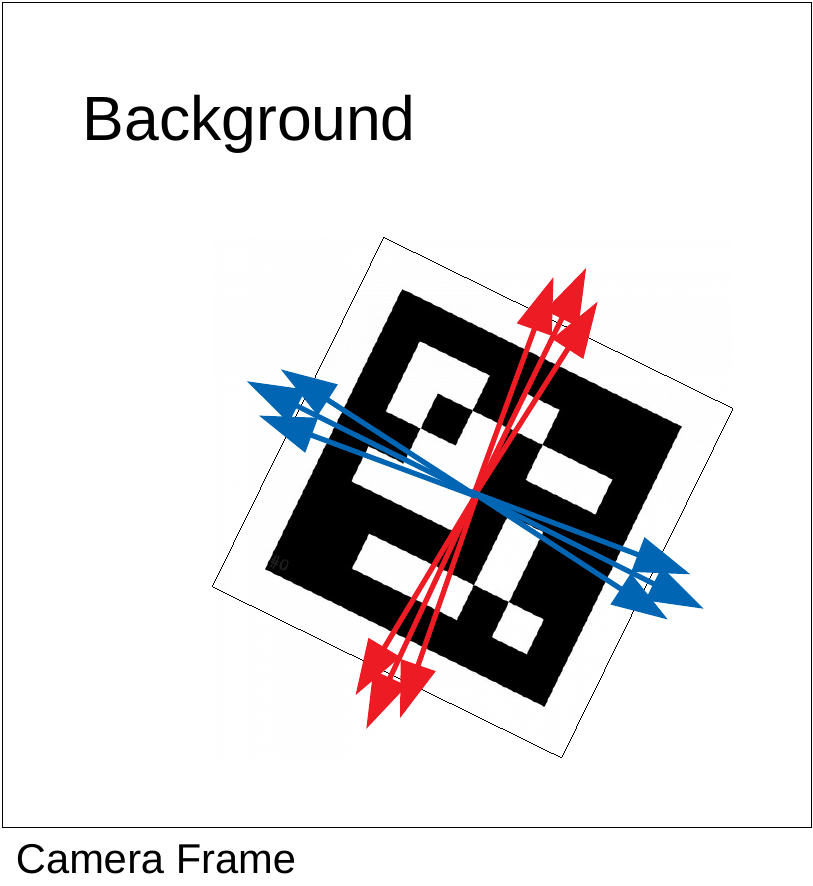}
    \caption{An example of the marker detectable by our algorithm in the camera frame. The marker is detectable when moving roughly parallel to its edges which is visualized by red and blue arrows on the marker above.}
    \label{fig:moving_dirs}
\end{figure}

\begin{figure*}[t]
    \centering
    \includegraphics[width=\textwidth]{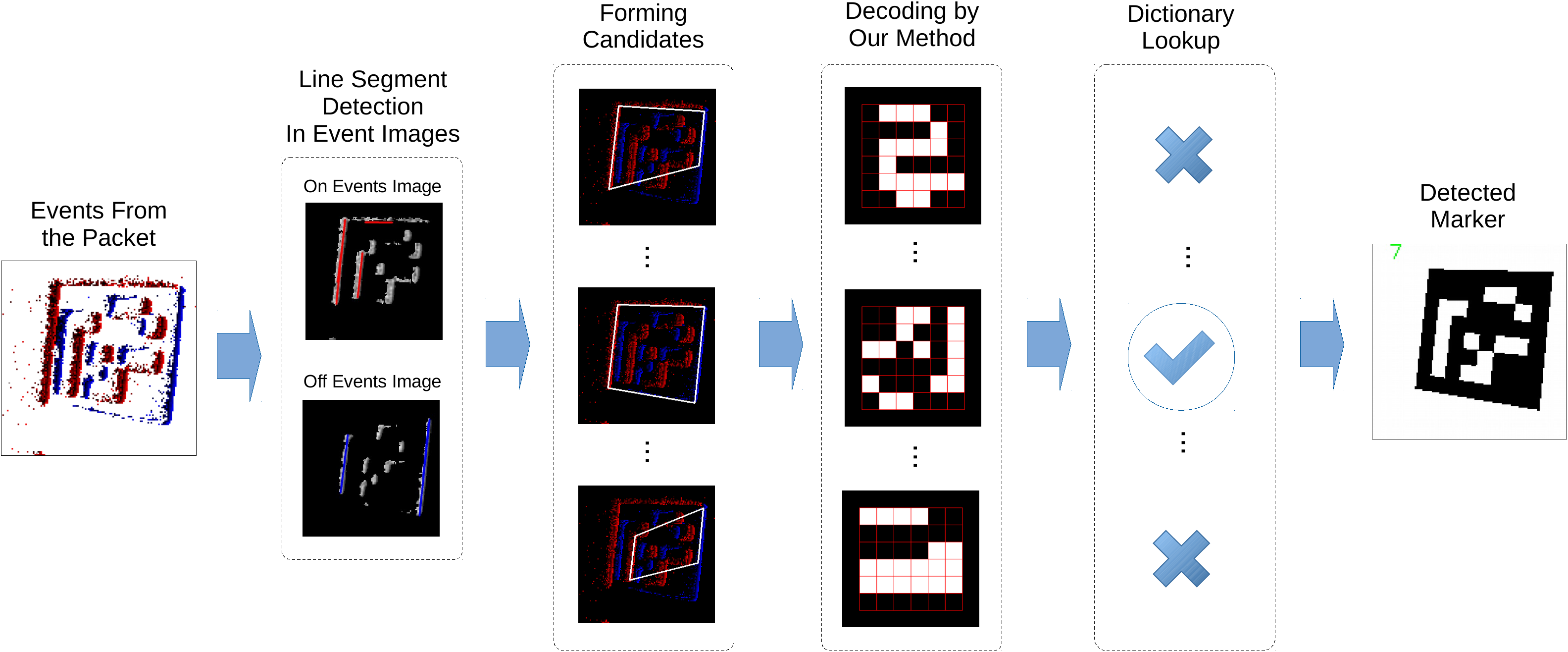}
    \caption{Overview of our marker detection algorithm from an event packet. Please note that the red color represents the \emph{on} events and line segment detections in the \emph{on} events image, and blue color represents the same for \emph{off} events.}
    \label{fig:overview}
\end{figure*}
\section{Methodology}
\label{sec::methodology}

We have developed our method using ArUco markers however it is applicable to all binary square markers since they all contain a binary grid of black and white cells surrounded by black borders. You can see an example of an ArUco marker and its cell grid in Figure \ref{fig:aruco_structure}. As can be observed, the cells on the edges of the grid are colored black and the marker is supposed to be printed on a white background to be easily detectable. The cells which are not on the border can be colored black or white. A binary code is associated to each marker by assigning 0 and 1 bits to black and white inner cells and forming a binary number by concatenating them. These binary codes come from a dictionary with codes that have a high Hamming distance from each other to reduce their chance of being confused. In the process of detection, the binary code is extracted from the marker and looked up in this dictionary to check if they are valid and determine their ID.

Our algorithm is capable of detecting markers that move in front of an event camera in the directions that are roughly parallel to its edges as visualized in Figure \ref{fig:moving_dirs}. Event cameras send events by packets that contain events from a fixed period of time, which is normally very short (e.g. 10 ms). You can find a visual overview of our approach in Figure \ref{fig:overview}. At the first step, events from each packet are converted into two separate images, one for \emph{on} and another for \emph{off} events. After doing some preprocessing on these images we detect line segments in them that correspond to the edges of the marker. Then we create marker candidates by combining single line segments from the corresponding \emph{on} and \emph{off} images. After that, the pixels within each candidate is unwarped to a standard square representation of the marker. We decode the marker by convolving Gaussian filters in positions where we expect events should exist to decode the color within each cell of the marker. Finally, the reconstructed colors are used to extract the binary code and it is compared to the codes within the marker dictionary to detect the ID number. 
\subsection{Creating Event Images from Event Packets}
\label{sec:event_image}

The procedure of creating an image from \emph{on} events and \emph{off} events is identical and is done separately for each event type. Hence, to avoid repetition, the procedure is described without specifying the event type.

Each event $e_i=(x,y,t)$ consists of three attributes, $x$ and $y$ are pixels coordinates and $t$ is the time stamp. Because the pixel at $(x,y)$ can have several events in the packet with different timestamps, $t$ is taken as the minimum timestamp among all events at $(x,y)$.

Let us assume that $E$ is the set containing all events of one type in the current event packet:
\begin{align}
E=\{e_i| i=1\ldots N\}
\end{align}
where $N$ is the number of events in $E$. We create an image $I$ with the same resolution as the event camera for each type of event. Hence we define the set of all pixel locations in the image by:
\begin{equation}
G=\{1\ldots W\}\times\{1\ldots H\}
\end{equation}
where $W\in \mathbb{Z}^+$ is the width and $H\in \mathbb{Z}^+$ is the height of the image in pixels.
We also define $S$ as the set of the pixel coordinates where at least one event exists:
\begin{align}
S=\{(x,y)\in G|\exists i \in \{1\ldots N\} :(x,y,t_i) \in E\}.
\end{align}
We call every pixel belonging to $S$ a \emph{valid} pixel and  put $$I(x,y)=t_i, \forall (x,y) \in S.$$

In order to simplify the timestamps, we normalize them according to the minimum and maximum timestamps in the packet and put the values in the image $I_{norm}$. We also set the non-valid pixels to zero:
\begin{align}
I_\text{norm}(x,y)=\begin{cases}
\frac{I(x,y)-t^\text{max}}{t^\text{max} - t^\text{min}},&\quad (x,y)\in S\\
0,&\quad (x,y)\in \overline{S}=G\setminus S
\end{cases}
\end{align}
where  $\setminus$ is the set difference operator and $\overline{S}$ is the set of non-valid pixels, and furthermore:
\begin{align}
    t^\text{min}=\min \{t_i\}_{i=1}^{N} \quad \text{and} \quad   t^\text{max}=\max \{t_i\}_{i=1}^{N}.
\end{align}

We were able to get better line segment detections (increase in marker detections by around 10\% of total detections) with $1-I_\text{norm}$ values as opposed to $I_\text{norm}$ values, hence we define:

\begin{equation}
    I'_\text{norm}(x,y)=
    \begin{cases}
    1 - I_\text{norm}(x,y), & (x,y) \in S\\
    0,& (x,y)\in \overline{S}
    \end{cases}
\end{equation}

Next, we fill the holes and refine the events in $I'_\text{norm}$ with the following function:
\begin{equation}
I_\text{ref}(x,y)=\begin{cases}
\frac{\sum\limits_{(x',y')\in B^S(x,y)}I'_\text{norm}(x',y')}{|B^S(x,y)|} & (x,y)\in F\\
&\\
0 & (x,y)\in F'\\
&\\
I'_\text{norm}(x,y)& 
(x,y) \in G \setminus (F \cup F')
\end{cases}
\end{equation}
where:
\begin{equation}
\begin{multlined}
    B(x,y)=\left\{(x',y')\in G \left|
    \begin{array}{c}
         x-1\leq x' \leq x+1\\
         y-1 \leq y' \leq y+1\\
    \end{array}
    \right.\right\}
\end{multlined}
\end{equation}
is the set of neighbors if pixels position $(x,y)$ in $G$,
\begin{equation}
B^S(x,y)=B(x,y)\cap S
\end{equation}
is the set of neighbors at $(x,y)$ which contain events,
\begin{equation}
    F= \left\{(x,y)\in \overline{S}\left| |B^S(x,y)|>\frac{|B(x,y)|}{2}\right.\right\}
\end{equation}
is the set of pixel positions with no event that most of their neighbors have events, and:
\begin{equation}
    F'=\left\{ (x,y)\in S \left| |B(x,y)\cap\overline{S}|>\frac{|B(x,y)|}{2}
    \right.\right\}
\end{equation}
is the set of positions with events that most of their neighbors have no events.
After that, we smooth the $I_\text{ref}$ image using a 2D Gaussian filter, $g_{n_s\times n_s}$ to obtain the $I_\text{smooth}$ image. Here, $n_s$ is the width and height of the Gaussian filter in pixels which also has the standard deviation of $\sigma_s$. The smoothing is done by the following function to reduce the noise from the sensor:
\begin{align}
\label{eq:smooth}
I_\text{smooth}(x,y)=
    \begin{cases}
    \frac{(I_\text{ref}*g)(x,y)}{(M*g)(x,y)}, & (x,y)\in S'\\
    0, & (x,y)\in G\setminus S'
    \end{cases}
\end{align}
Here, $S'=(S\cup F)\setminus F'$, which is the set of pixels with valid refined events and $M$ is an image mask define by:

\begin{align}
    M(x,y)=\left\{ \begin{array}{ll}
        1 & (x,y) \in S' \\
        0 & G\setminus S'
    \end{array} \right..
\end{align}
In Equation \ref{eq:smooth}, we divide the convolution of the normalized image by the convolution of the mask to eliminate the effect of non-valid pixels in smoothing the valid pixel values.

As said before, the procedure mentioned so far is applied separately to \textit{on} and \textit{off} events to obtain \emph{two} images. The next step is to detect line segments in these images.

\subsection{Detection of Line Segments in Event Images}
Since binary square markers have a rectangular black margin on a white background it is possible to detect their edges in event images when they are moving. A line in \emph{off} events image will appear on the edge on the direction of movement, and a line of \emph{on} events will appear on the edge on the opposite direction of movement.
We take advantage of the LSD line segment detector introduced in \cite{gioi_lsd_2012}. We chose this detector because it is faster and produces fewer and more accurate candidate than other methods such as the Hough Transformation \cite{duda_use_1972}. We remove line segments that are shorter than a minimum length, $\text{l}_\text{min}$, from the output of the LSD algorithm. This helps to remove candidates that are too far away from the camera or have too much of the perspective effect to be decoded.

Let us take $L$ as the set of all detected valid line segments:
\begin{align}
    L= \left \{ l_i \right \}_{i=1}^{N_L},\quad l_i=\big((x_1^i,y_1^i),(x_2^i,y_2^i) \big)
\end{align}
Here, $(x_1^i,y_1^i)\in \mathbb{R}^2$ and $(x_2^i,y_2^i)\in \mathbb{R}^2$ are the two ends of the line segment $l_i$ and are positioned within the boundaries of the camera's image resolution. We also define the function:
\begin{equation}
\begin{multlined}
    \text{length}(l_i)=||(x_1^i-x_2^i,y_1^i-y_2^i)||_2,\\ i=1\ldots N_L
\end{multlined}
\end{equation}
which returns the length of the line segment $l_i$.

\subsection{Line Segments' Age Correction}
In order to have segments that correspond to the same time in the packet's time interval, we correct the position of the line segments so that their occurrence falls in the middle of the time interval. For this purpose we define a measurement of their \emph{age} which is the average normalized timestamp of the pixels that fall on the line segment. More formally the age of the line segment $l=((x_1,y_1),(x_2,y_2))$ is obtained as:
\begin{align}
    \text{age}(l)=\bar a,\quad \text{for} \quad \text{all} \quad a\in A(l)
\end{align}
where
\begin{align}
    A(l) = \left\{ I_\text{norm}(x,y)  \left| 
    \begin{array}{c}
        M(x,y)=1\; \\
        (x,y)\in P(l)
    \end{array}\right. \right\}
\end{align}
and
\begin{align}
    P(l)= \left\{ (x,y) \left | 
    \begin{array}{c}
    x=\lfloor s_x+x_1+0.5  \rfloor \land \\
         y=\lfloor s_y+y_1+0.5\rfloor \land\\
         0\geq s_x \geq \lfloor x_2-x_1+0.5\rfloor \land\\
         0\geq s_y \geq \lfloor y_2-y_1+0.5\rfloor
    \end{array}\right. \right\}.
\end{align}
Here $P(l)$ is the set of all pixels on the line segment $l$, $A(l)$ is the set of pixel values on the segment in image $I_\text{norm}$, $\bar a$ is the average value of all elements in $A(l)$, and $s_x$ and $s_y$ are the steps in $x$ and $y$ directions for moving on the line segment.

To make our marker decoding more robust we would like our line segments to contain events that are in the middle of the packet's time interval, i.e., we want the age of the line segments to be equal to 0.5. However, our line segment detector does not always return segments with such age. Hence we move the line segment perpendicular to its orientation in two directions until we find the position with the right age (0.5). The formal definition of this algorithm is presented in Algorithm \ref{alg:line_age}.
\begin{algorithm}
\caption{Line Age Correction}
\label{alg:line_age}
\begin{algorithmic}
\Procedure{CorrectAge}{$l=\big((x_1,y_1),(x_2,y_2)\big)$}
\State{$\vec u \gets (y_2-y_1,x_1-x_2)$} 
\State{$\vec u \gets \vec u/||\vec u||_2$} \Comment{$\vec u$: unit vector perpendicular to $l$}
\State{$R \gets \left\{\big (0,\text{age}(l) \big)\right\}$}
\For{$d\in\{-1,1\}$}
\State{$\vec v\gets d\vec u$}
\State{$s\gets1$}
\State{$l_\text{new} \gets \left(s \vec v+(x_1,y_1),s \vec v+(x_2,y_2)\right)$}
\While{$|A(l_\text{new})|<|P(l_\text{new})|/2$}
\State{$R \gets R \cup \{ (sd,\text{age}(l_\text{new})) \}$}

\State{$s\gets s+1$}
\State{$l_\text{new} \gets \left(s \vec v+(x_1,y_1),s \vec v+(x_2,y_2)\right)$}
\EndWhile
\EndFor
\State{$(\hat\alpha,\hat\beta)=\arg \min\limits_{\alpha,\beta} \left (\sum\limits_{(x,y)\in R} (\alpha x + \beta-y)^2 \right )$}
\State{}\Comment{$\alpha,\beta$ are parameters for linear regression over the tuples in $R$}
\State{$t\gets (0.5-\hat\beta)/\hat\alpha$}
\State{\Return $(t\vec u+(x_1,y_1),t \vec u+(x_2,y_2))$}
\EndProcedure
\end{algorithmic}
\end{algorithm}

We correct the age of all detected line segments. Hence we define:
\begin{align}
    L^\text{corrected}=\{\textsc{CorrectAge}(l_i)\}_{i=1}^{N_L}
\end{align}

The next step is to use the line segments in $L^\text{corrected}$ to form candidates for marker detection.
\begin{figure*}[t!]
    \centering
    \includegraphics[width=\textwidth]{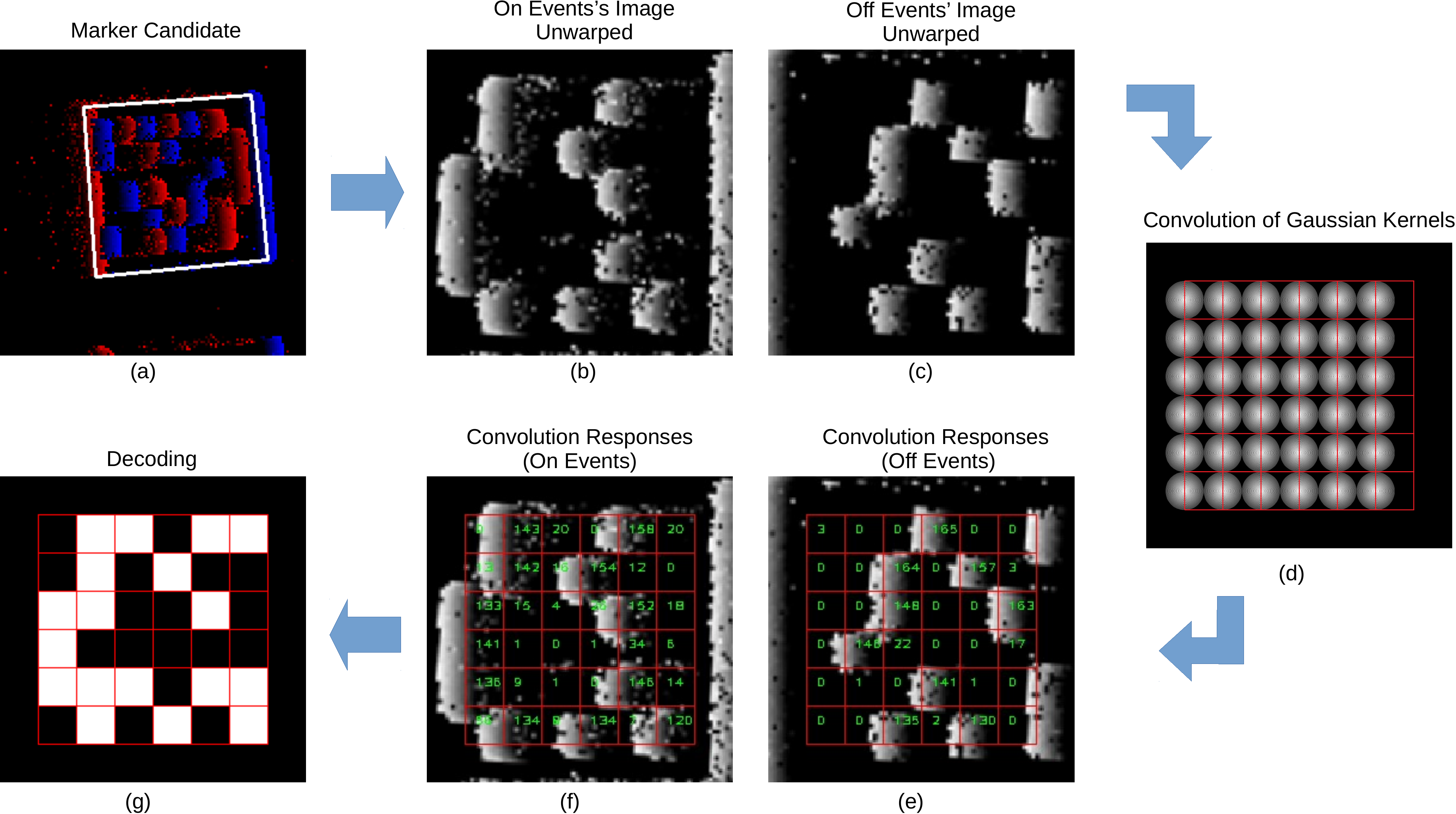}
    \caption{Summary of our marker detection algorithm. First the marker candidate in (a) is unwarped to a standard square images related to \emph{on} (b) and \emph{off} (c) events. Then Gaussian kernels on a grid (d) are convolved with the images and the responses at convolution points are saved for \emph{on} (f) and \emph{off} (e) events. Finally, the responses are used to reconstruct the marker (g).}
    \label{fig:decoding}
\end{figure*}

\subsection{Creation of Marker Candidates}

In order to detect candidates for marker detection we need to match line segments from \emph{on} events to the ones from \emph{off} events. We make some conditions to make it more likely for the line segments to represent a valid marker candidate. Let us assume that $L^\text{corrected}_\text{on}$ and $L^\text{corrected}_\text{off}$ are the corrected line segments related to \emph{on} and \emph{off} events respectively such that:
\begin{equation}
    L^\text{corrected}_\text{off}=\{l^\text{off}_i\}_\text{i=1} ^{N_{L}^\text{off}}, L^\text{corrected}_\text{on}=\{l^\text{on}_i\}_\text{i=1} ^{N_{L}^\text{on}}
\end{equation}
Then we form the set of our marker candidates, $C$, in the following way:
\begin{equation}
\begin{multlined}
    C=\\\left\{(l_i^\text{on},l_j^\text{off}) \left |
    \begin{array}{c}
    \text{length}(l_i^\text{on})< 2\times\text{length}(l_j^\text{off}) \land \\
    \text{length}(l_j^\text{off})< 2\times\text{length}(l_i^\text{on}) \land \\
     \left[\text{Project}(l_i^\text{on},l_j^\text{off}) \lor
    \text{Project}(l_j^\text{off},l_i^\text{on}) \right] \land \\ 
    \gamma(l_i^\text{on},l_i^\text{off})\leq \frac{\pi}{6}
    \end{array}
    \right . \right\}
\end{multlined}
\end{equation}
where $\gamma(l_i^\text{on},l_i^\text{off})$ is the minimum angle between the two line segments $l_i^\text{on}$ and $l_i^\text{off}$, and $\text{Project}( ., .)$ is the function that takes two line segments and makes sure if at least one of the end points of one segment projects within the end points of the other line segment:
\begin{equation}
\begin{multlined}
    \text{Project}((\vec b_1, \vec e_1), (\vec b_2, \vec e_2))=\\0\leq\frac{(\vec e_1-\vec b_1).\vec b_2}{||e_1-b_1||_2}\leq||e_1-b_1||_2\quad \lor\\
    0\leq\frac{(\vec e_1-\vec b_1).\vec e_2}{||e_1-b_1||_2}\leq||e_1-b_1||_2
\end{multlined}
\end{equation}
Here $(\vec b_1,\vec e_1)$ and $(\vec b_2,\vec e_2)$ are, respectively, the coordinate vector pairs of the beginning and end points for the first and the second input line segments to $\text{Project}(.,.)$ .
Now that we have formed the candidates for marker detection, we will attempt to decode them.
\subsection{Decoding of Marker Candidates}
\begin{figure*}
    \centering
    \includegraphics[width=\textwidth]{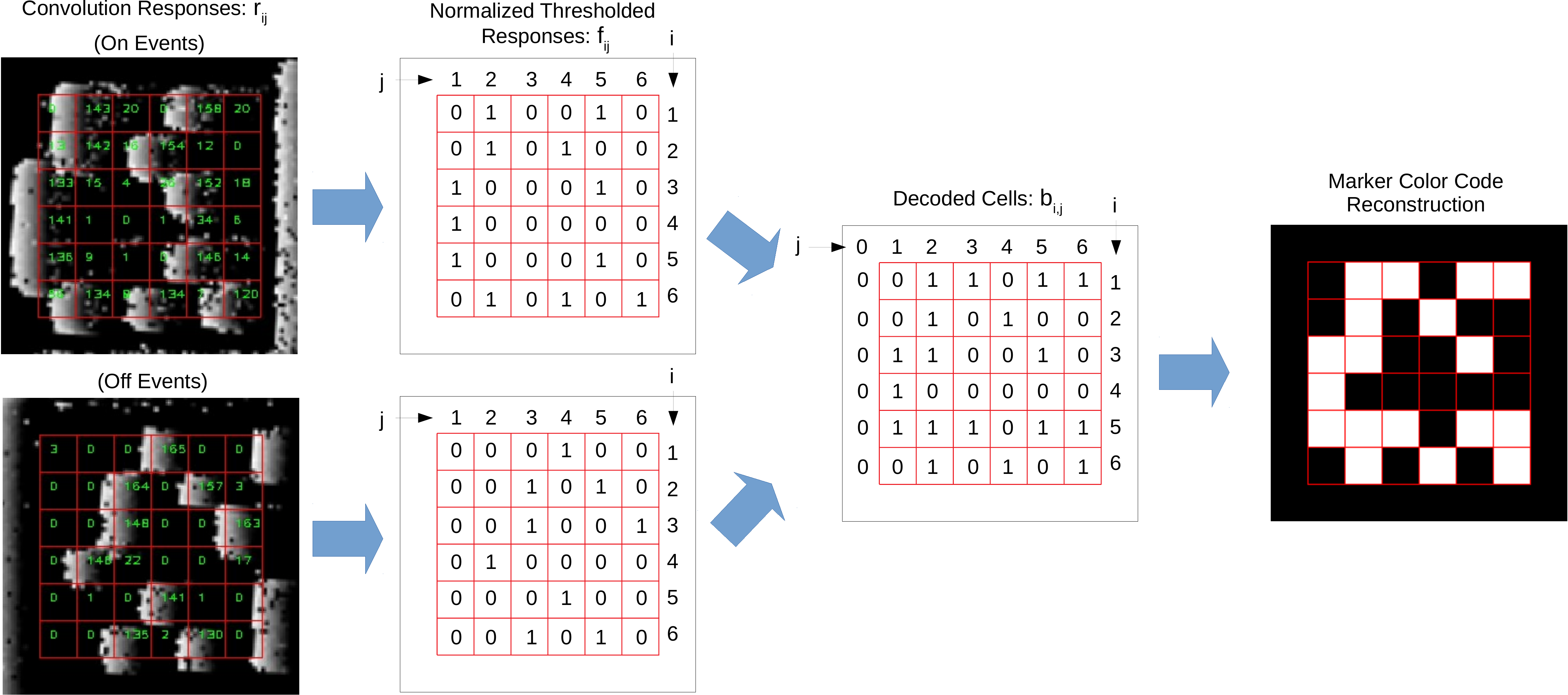}
    \caption{Details of decoding the markers from Gaussian convolution responses. From left to right: First responses of Gaussian convolutions are stored for each code cell of the marker. Then the response values are normalized according to the highest response and then thresholded. After that, cells at each row of the marker are decoded according to the normalized thresholded responses of \emph{on} and \emph{off} events and decoded value of the cell on the left. Finally, the decoded marker is decoded and can be reconstructed. }
    \label{fig:decoding_detail}
\end{figure*}

The first step in decoding a marker candidate is to unwarp it to a square image with standard dimensions. To decode the candidates we use the image $I_\text{norm}$ which was defined in Section \ref{sec:event_image}.

To unwarp the candidates for decoding, we find the perspective transformation that maps the corners of the marker in the $I_\text{norm}$ image to the corners of the target image with standard dimensions. We denote the unwarped image for the $i$-th candidate by $I_C^i$:
\begin{equation}
    I_C^i=\text{UnwarpPerspective}(c_i,s_c,I_\text{norm}),\quad i=1\ldots N_C
\end{equation}
Here, $s_c$ is the dimension of the standard square image in pixels, $c_i$ is the $i$-th marker candidate from $C$, and $N_C$ is the number of marker candidates, hence:
\begin{equation}
   C= \{c_i\}_{i=1}^{N_C}.
\end{equation}
The unwarping is done in a manner that the edge detected by the line segment from the \emph{off} events is always on the left and the one from \emph{on} events is on the right. Therefore the orientation of the unwarped marker would be as if it is always moving towards the left direction.

You can see an example of unwarping the candidate from $I_{norm}$ in Figure \ref{fig:decoding}(b) and \ref{fig:decoding}(c) which are related to \emph{on} and \emph{off} events respectively. The polygon representing the candidate is shown by white borders in Figure \ref{fig:decoding}(a) where the $I_{norm}$ image of \emph{on} events (in red) and the $I_{norm}$ image of \emph{off} events (in blue) are shown overlaid on top of each other.

After obtaining the unwarped standard candidate images, they are segmented into square regions (inner cells) according to ArUco's specification. For each of these cells, we need to determine if the color inside is white or black. Since in the standard unwarped images the marker is always supposed to move to the left, we check the events on the left side of each cell. This is done by convolving a 2D Gaussian kernel (with the standard deviation of $\sigma_d$) on each square cell's left edge. The grid used for decoding along with the positions where the Gaussians are convolved with the unwarped images is shown in Figure \ref{fig:decoding}(d) where you can see a visualization of the Gaussians on the left edge of the cells.
Although, in theory, the events should appear exactly on the edges of the cell squares on the marker, in practice, because the \emph{on} and \emph{off} events are not perfectly synchronized, this might not be the case. For example in Figure \ref{fig:decoding}(f) you can see that the \emph{on} events on the edge of each square is slightly shifted to the left while for \emph{off} events (Figure \ref{fig:decoding}(e)) this issue does not exists. For this reason, we have to shift the convolution point of the Gaussian kernel slightly to the left for \emph{off} pixels. We set the amount of shifting to $n_d/4$ pixels where $n_d$ is the side length of each cell square (and also the Gaussian kernel) in pixels.
When the convolutions are applied, they result in values used to determine the color of each cell of the marker. The scores within each cell are shown in Figure \ref{fig:decoding}(e) and \ref{fig:decoding}(f). 

Let us assume that $r_{ij}$ is the response of the convolution related to the inner cell on the $i$-th row and $j$-th column on the marker. We normalize these responses according to the maximum response and then threshold them:
\begin{equation}
    f_{ij}=\left\lfloor(\frac{r_{ij}}{\max\limits_{i,j}r_{ij}})/\theta\right\rfloor,\quad i,j \in \{1\ldots N_m\} 
\end{equation}
where $N_m$ is the size of the code square of the ArUco marker, $\theta$ is the threshold for the cell response, and $f_{ij}$ determines if the events occur on the left edge of the cell. We obtain this value separately for \emph{on} and \emph{off} events, hence, we have both $f_{ij}^\text{on}$ and $f_{ij}^\text{off}$ for each cell at position $(i,j)$. Now we can determine the color code within each cell of the marker as follows:

\begin{multline}
    b_{i,j}=\begin{cases}
    0 & \quad j=0 \\
    &\\
    1-b_{i,j-1} & \quad \begin{array}{l}(b_{i,j-1}=0 \land f_{ij}^\text{on}=1) \\
    \lor (b_{i,j-1}=1 \land f_{ij}^\text{off}=1) \end{array}\\
    &\\
    b_{i,j-1} & \quad otherwise \end{cases}
\end{multline}
where $b_{i,j}, \forall i=1\ldots N_m,\; j=0\ldots N_m$ determines the color code at cell $(i,j)$ in the marker. In Figure \ref{fig:decoding_detail} you can find $r_{i,j}$, $f_{i,j}$, and $b_{i,j}$ values for an example of a marker that is successfully reconstructed.
In the last stage, the binary code is extracted from the reconstructed marker and checked for its validity and its ID according to the marker specifications. In the case of ArUco markers, this is done by concatenating each row of cell codes and looking up the resulting binary number in the marker dictionary. This is done four times for each rotation of the reconstructed marker. If the code is not found in the dictionary we reject it as a false candidate, otherwise, its ID is extracted from the dictionary.
\begin{figure*}[t!]
    \centering
    \subcaptionbox{\label{fig:cam_rig}}{\includegraphics[width=.48\textwidth]{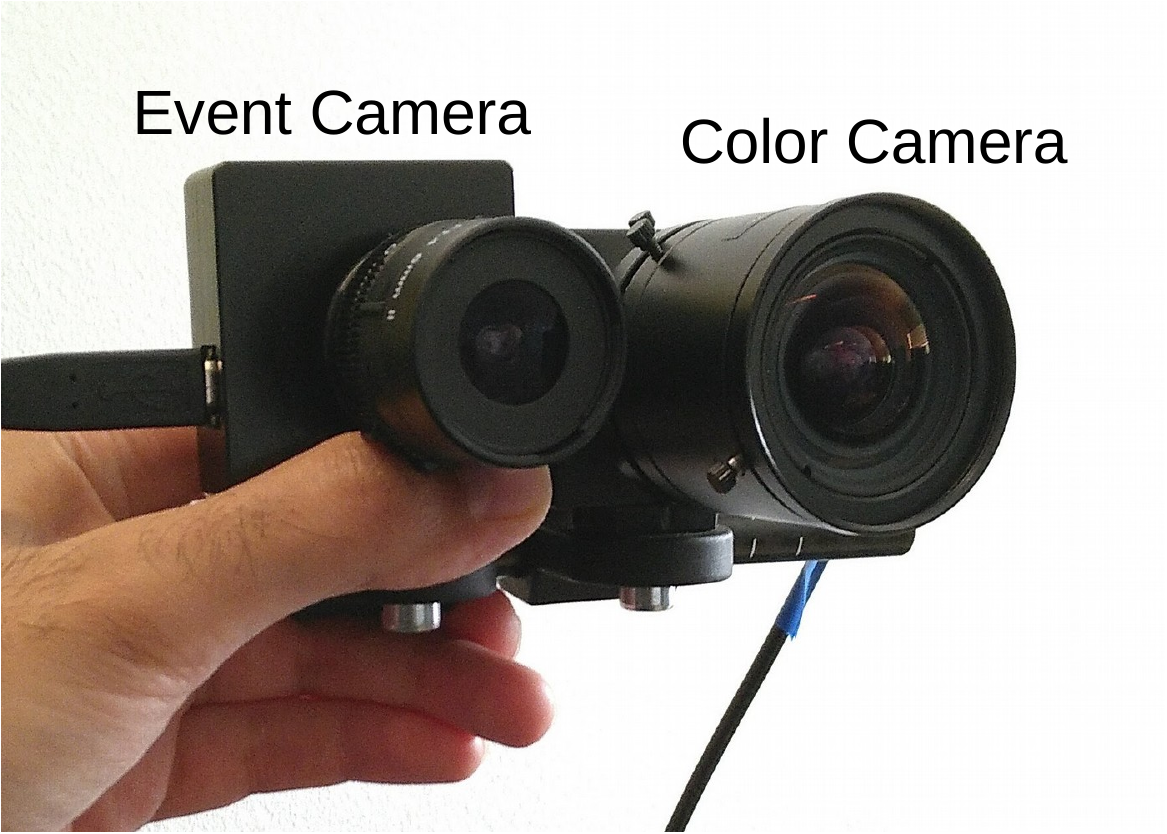}}
    \subcaptionbox{\label{fig:marker_grid}}{\includegraphics[width=.48\textwidth]{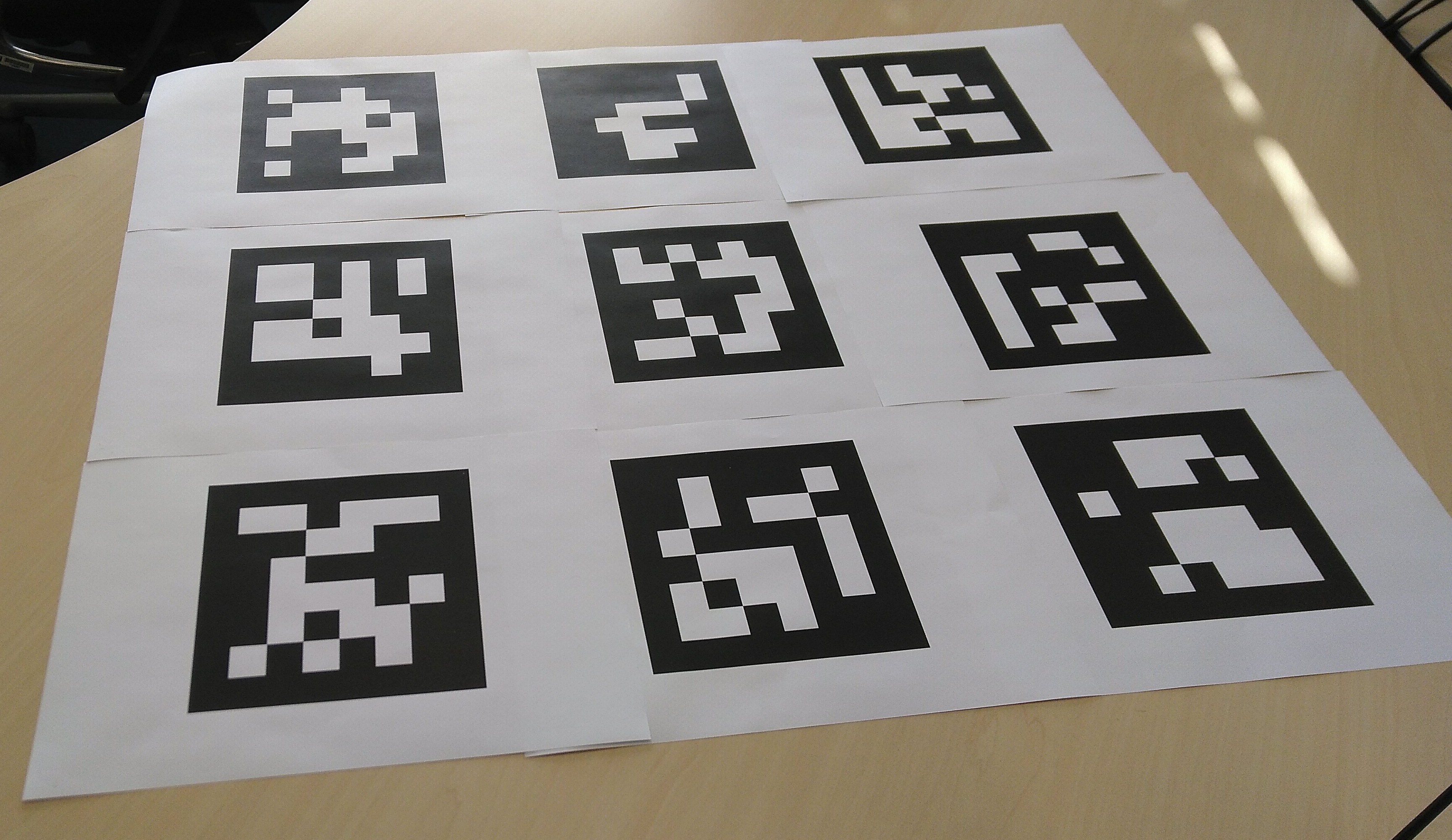}}
    \caption{Our experimental setup including (\subref{fig:cam_rig}) the camera rig we employed for capturing our dataset and (\subref{fig:marker_grid}) the $9\times9$ marker grid we used for evaluation of our algorithm. The color and the event camera are fixed on the same rod facing the same direction. The sequences were captured moving the camera rig by hand in left-right and up-down directions while pointing perpendicularly at the marker grid.}
    \label{fig:setup}
\end{figure*}

\section{Experimental Results}
\label{sec::experiments}
To test our algorithm we captured some sequences of moving the event camera in front of ArUco markers. These sequences were recorded\footnote{The recording was done in the Interdisciplinary Centre for Security, Reliability and Trust (SnT) at University of Luxembourg} using the iniVation DVS128 event camera, with a resolution of $128\times128$ pixels.
The event camera was attached to a camera rig together with a color camera to verify the current scene in the color image as well. The color camera was an IDS UI-1220LE-C-HQ with a global shutter and a resolution of $752 \times 480$ pixels at $60$ fps. Figure \ref{fig:setup}(\subref{fig:cam_rig}) shows  our camera rig setup. We ran our algorithm on a laptop with the Core™ i7-4700HQ CPU and 8 GB of RAM. We implemented our algorithm in C++ and did our experiments under the Ubuntu 18.04 operating system. Our implementation only needs a single CPU core to run. To reduce the noise from the camera to some extent we employed the DVS noise filter from the libcaer library\footnote{https://gitlab.com/inivation/dv/libcaer}.

We did both quantitative and qualitative evaluations of our algorithm using the data we captured from the event and color camera. To do so we captured sequences by holding the camera close to a grid of markers and moving it in different directions with respect to the markers while pointing perpendicularly at them. The recording was done in an environment with controlled lighting and low ambient light noise. You can see an image of the marker grid in Figure \ref{fig:setup}(\subref{fig:marker_grid}). Because of the low resolution of the event camera, we had to keep the cameras close to the markers, at approximately $20$ to $30$ cm away from them. In this way, only one marker was visible at a time in the field of view of both cameras.

The captured sequences contain $3470$ packets in total where each packet contains the events from a $10$ ms time interval. We aggregated all packets and color frames from the sequences together for our quantitative evaluation. For a minority number of packets ($<100$), there were distortions in the events because of the bandwidth overflow, we discarded these packets from our evaluations.

\begin{table}[]
    \centering
    \begin{tabular}{|c|c|c|}\hline
    Symbol & Value & Description \\\hline\hline
    $W$ & 128 & Event image width\\\hline
    $H$ & 128 & Event image height\\\hline
    $n_s$ & 3 & Gaussian filter size for $I_\text{smooth}$\\\hline
    $\sigma_s$ & $0.8$& Gaussian filter sigma for $I_\text{smooth}$\\\hline
    $\text{l}_\text{min}$ &25& Minimum line segment size\\\hline
    $s_c$ &160& Standard marker image size\\\hline
    $n_d$ &20& Cells size for marker image\\\hline
    $\sigma_d$ &3.35& Gaussian filter sigma for decoding\\\hline
    $\theta$ & 0.55 & Threshold for cell responses\\\hline
    \end{tabular}
    \caption{Parameters we have used for our evaluations of our approach. The unit for all parameters except $\theta$ (which has no unit) is pixels.}
    \label{tab:parameters}
\end{table}
The parameter values we have employed for our algorithm during our experiments are shown in Table \ref{tab:parameters}.
\subsection{Running Speed}
We ran our implementation on the captured sequences and measured the time needed for processing each packet. Then we calculated the average of the durations which turned out to be $8.44$ milliseconds per packet on a single CPU core. This means that our implementation can handle $118$ packets per second which are already higher than the $100$ packets that our camera produces in each second. Hence we were able to run our algorithm in real-time. For further analysis, we also calculated the amount of time needed for different steps of the algorithm. The values are presented in Table \ref{tab:relative_time} in milliseconds. As you can see the most time-consuming part of the algorithm is unwarping the candidates to the standard representation which takes $3.42\pm3.51$ ms. It also has a relatively high standard deviation value, the reason is that in some frames many marker candidates are formed while in others, due to less detected line segments, fewer candidates are generated which creates the variance in time needed for unwarping. After candidate unwarping, the LSD line segment detection and candidate formation takes most of the time with the average of $2.69\pm0.73$ ms.
\begin{table}[]
    \centering
    \begin{tabular}{|c | c|}\hline
        Step of Algorithm & Processing Time Per Packet \\\hline\hline
        \multirow{2}{3.8cm}{\centering Event Image Creation and the rest}& \multirow{2}{4cm}{\centering $2.13\pm0.69$ ms}\\
        & \\\hline
        \multirow{2}{3.8cm}{\centering Segment Detection and Candidate Formation}  & \multirow{2}{4cm}{\centering $2.69\pm0.73$ ms} \\
        & \\\hline
        Candidate Unwarping & $3.42\pm3.51$ ms\\\hline
        \multirow{2}{3.8cm}{\centering Marker Decoding and Code Look-up}  & \multirow{2}{4cm}{\centering $0.20\pm0.47$ ms}\\
        &\\
        \hline\hline
        Total & $8.44\pm4.48$ ms\\
        \hline
        
    \end{tabular}
    \caption{The average processing time and standard deviation for different steps of our algorithm and also in total for processing each packet (in milliseconds).}
    \label{tab:relative_time}
\end{table}
\subsection{Quantitative Evaluation and Discussion}
In our evaluation sequences, we moved the camera rig in side-to-side and up-and-down motions in front of the marker grid. We analyzed the sequences manually to separate the frames (packets) where a marker is completely visible in the event camera.

We measured two metrics. First, the percentage of times a marker passes in front of the camera and is detected. And second, the number of frames in which the marker is detected while visible in the event camera's frame. More precisely, we define that a marker ``passes'' in front of a camera when the whole marker appears in the field of view of the camera and moves in front of it until a part of it exits the field of view. If the marker is detected at any time while ``passing'', we say that we have a ``pass detection''. Likewise, frame detections are counted within frames where the whole marker is within the field of view of the camera.

Table \ref{tab:all_frames} presents the results related to pass detection rate and frame detection rate for the color camera (using ArUco algorithm) and for the event camera (using our algorithm). As can be observed our method has a very high pass detection rate ($94$\%) especially compared to that of ArUco's (3\%). The low frame detection and pass detection rate of ArUco suggests the high sensitivity of the ArUco algorithm to motion blur. On the other hand, it can be seen that our approach has detected $44$\% of the detectable frames, while ArUco only is able to detect $2$\%. 

We believe that the missed detections of our method are due to several reasons. First, since the camera rig is moved by hand in a lateral manner, it needs to stop at some points and change direction. In these frames, the movement slows down for a moment and then changes direction which causes few events to be created, and hence our algorithm would not perform well. Another reason is that at some frames the camera was not moving parallel enough to the edges of the marker (as it should similar to Figure \ref{fig:moving_dirs}). In these cases, our algorithm could not perform correctly, however, this is a limitation of the design.

\begin{table}[]
    \centering
    \begin{tabular}{|c|c|c|c|}\hline
         \multicolumn{4}{|c|}{Accuracy} \\
         \multicolumn{2}{|c|}{Frame Detection(\%)} & \multicolumn{2}{c|}{Pass Detection(\%)}\\ 
         ArUco & Ours & ArUco & Ours\\
        \hline\hline
         1.58 & 44.16 & 3.39 & 93.63 \\\hline
         
    \end{tabular}
    \caption{Accuracy of detected passes and detected frames when the whole marker is in the field of view. This is calculated independently for the color camera using ArUco algorithm and the event camera using our algorithm.}
    \label{tab:all_frames}
\end{table}

In order to make a deeper comparison, we also calculated the frame detection rate only for frames where the whole marker is in the field of view of both cameras. We also counted the frames where both ArUco and our algorithm detect the marker in the image where there was a ``mutual'' detection between the ArUco algorithm and our algorithm. The results are shown in Table \ref{tab:mutual_frames}. As you can see there were no frames that were mutually detected by both of the cameras. This indicates that our algorithm is a very good complement for the normal ArUco detection method. On the other hand, our detection rate ($35$\%) is still more than an order of magnitude better than that of ArUco's ($2$\%) which suggests the extreme sensitivity of the ArUco algorithm to motion blur and that our algorithm handles moving markers much better.

\begin{table}[]
    \centering
    \begin{tabular}{|c|c|c|}\hline
         \multicolumn{3}{|c|}{\multirow{2}{5cm}{\centering Accuracy of Frame Detection in Common Frames (\%)}} \\
         \multicolumn{3}{|c|}{} \\
         Mutual & ArUco & Ours \\ 
        \hline\hline
        0.00 & 1.67  &  35.34 \\\hline
    \end{tabular}
    \caption{Accuracy of detected frames when the marker is in the field of view of both cameras.}
    \label{tab:mutual_frames}
\end{table}

\begin{figure*}
\centering
\raisebox{1.5cm}{1  } \raisebox{.5cm}{\includegraphics[width=0.19\textwidth]{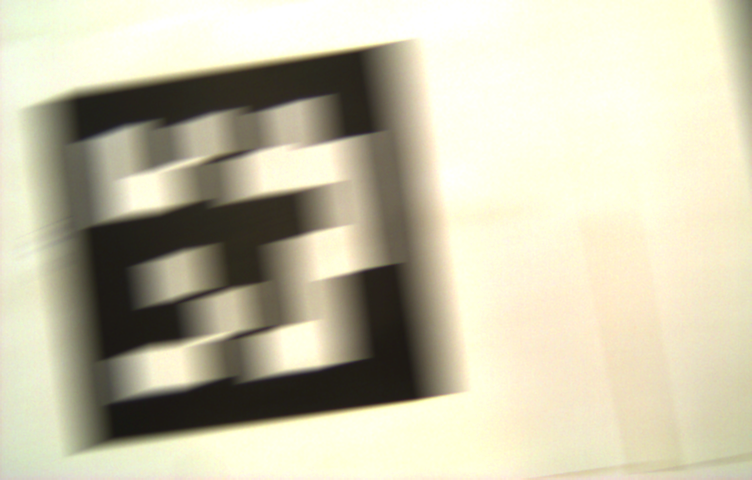}} \includegraphics[width=0.17\textwidth]{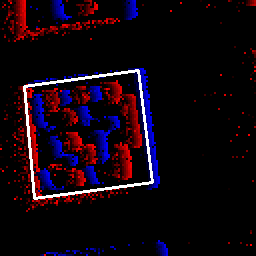}
\includegraphics[width=0.17\textwidth]{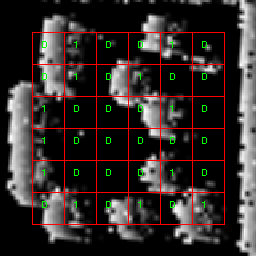}
\includegraphics[width=0.17\textwidth]{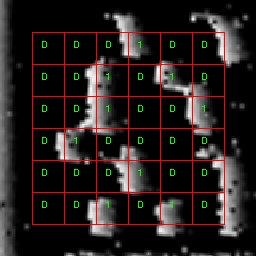}
\includegraphics[width=0.17\textwidth,frame]{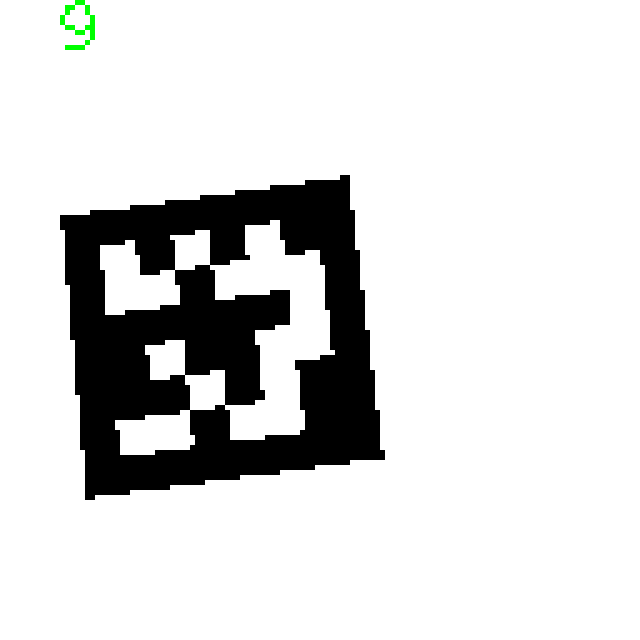}
\\
\raisebox{1.5cm}{2  } \raisebox{.5cm}{\includegraphics[width=0.19\textwidth]{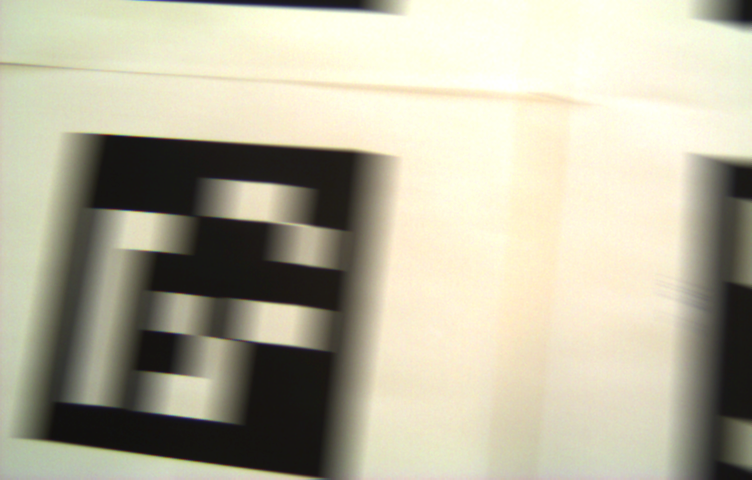}}
\includegraphics[width=0.17\textwidth]{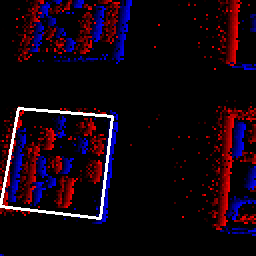}
\includegraphics[width=0.17\textwidth]{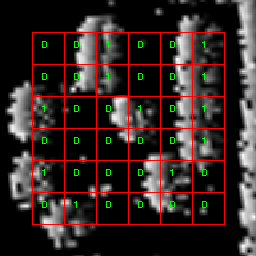}
\includegraphics[width=0.17\textwidth]{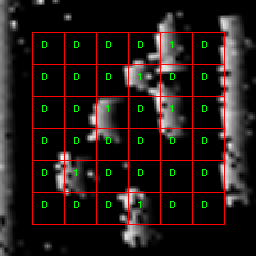}
\includegraphics[width=0.17\textwidth,frame]{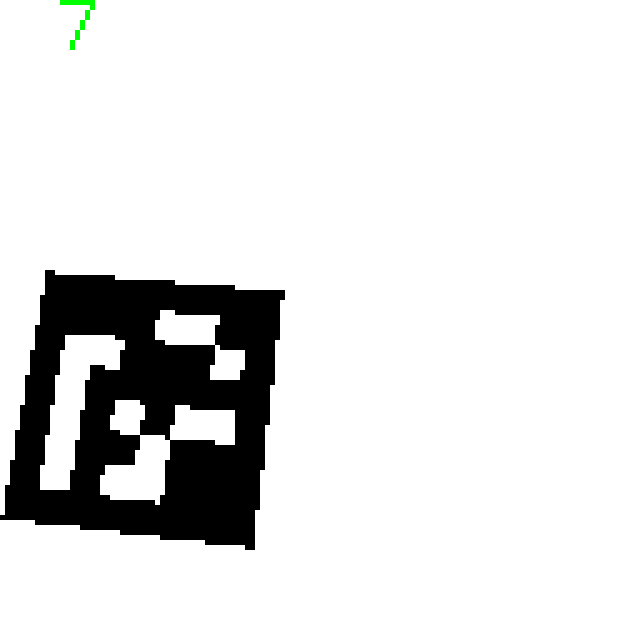}
\\
\raisebox{1.5cm}{3  } \raisebox{.5cm}{\includegraphics[width=0.19\textwidth]{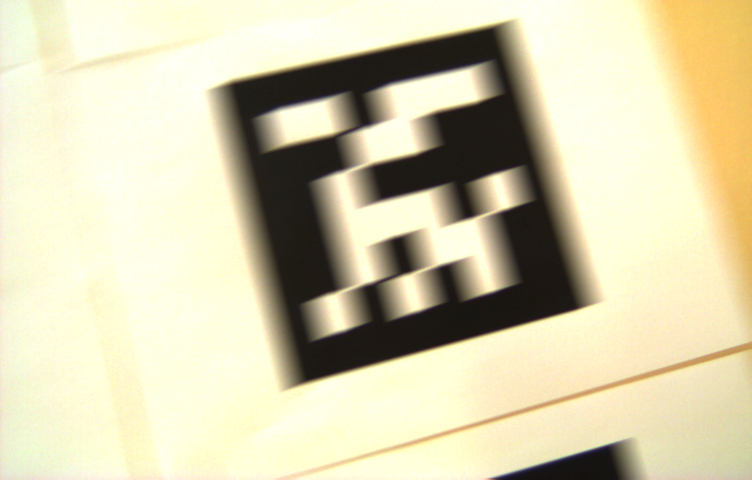}}
\includegraphics[width=0.17\textwidth]{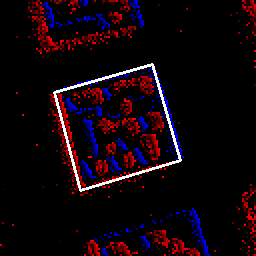}
\includegraphics[width=0.17\textwidth]{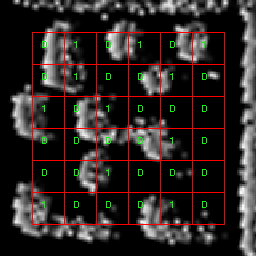}
\includegraphics[width=0.17\textwidth]{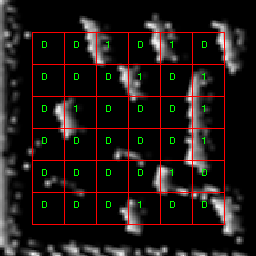}
\includegraphics[width=0.17\textwidth,frame]{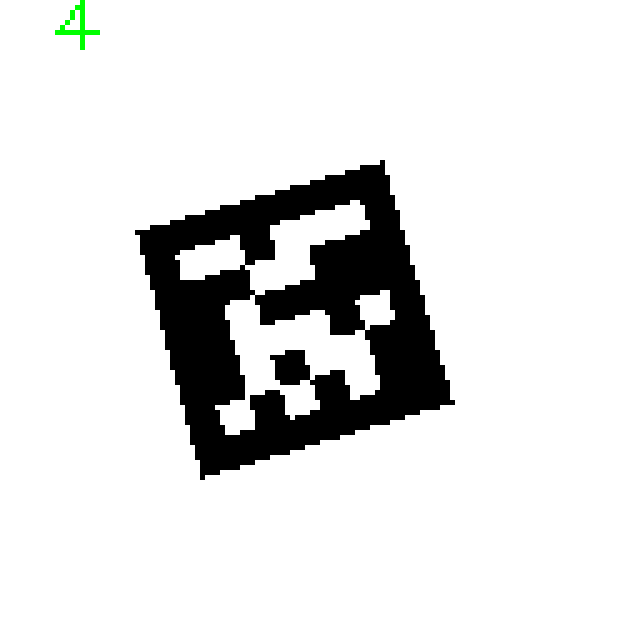}
\\
\raisebox{1.5cm}{4  } \raisebox{.5cm}{\includegraphics[width=0.19\textwidth]{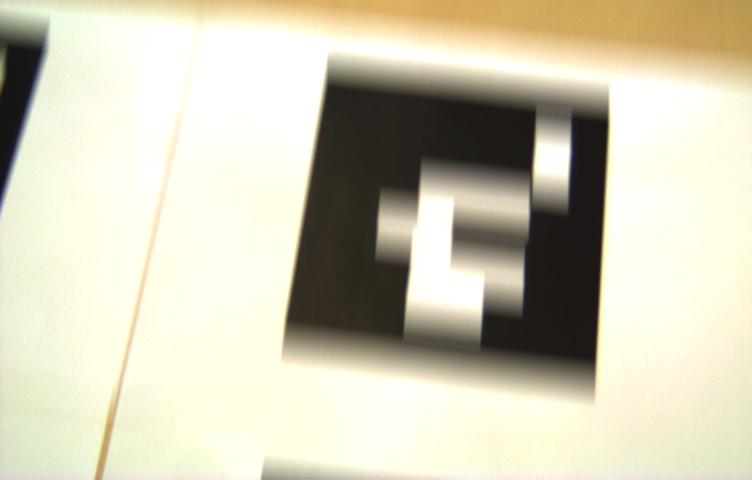}}
\includegraphics[width=0.17\textwidth]{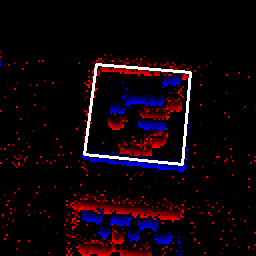}
\includegraphics[width=0.17\textwidth]{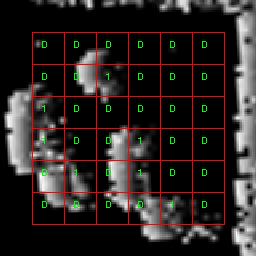}
\includegraphics[width=0.17\textwidth]{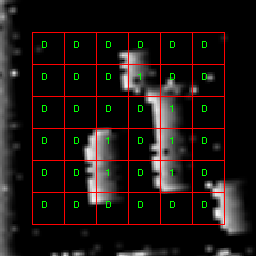}
\includegraphics[width=0.17\textwidth,frame]{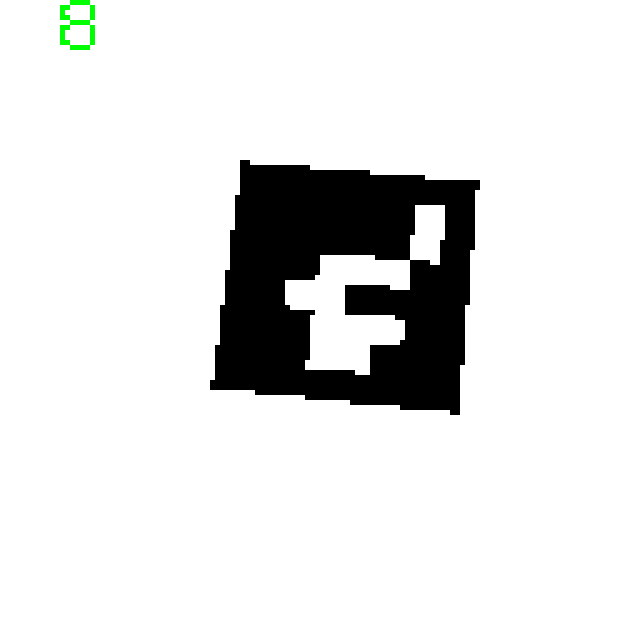}
\\
\raisebox{1.5cm}{5  } \raisebox{.5cm}{\includegraphics[width=0.19\textwidth]{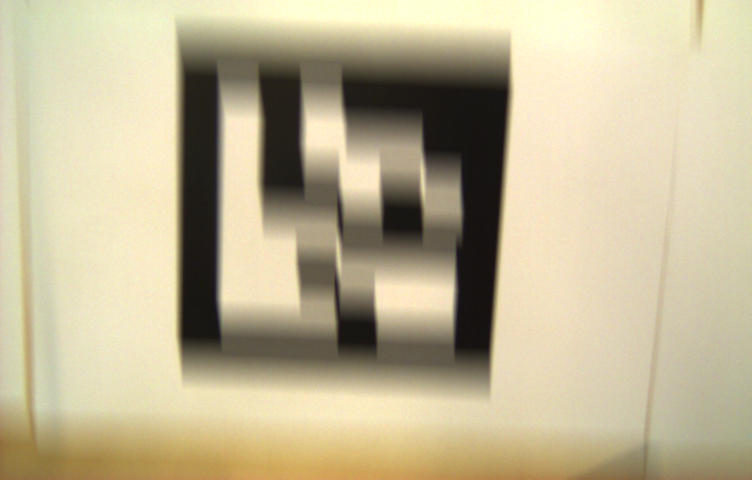}}
\includegraphics[width=0.17\textwidth]{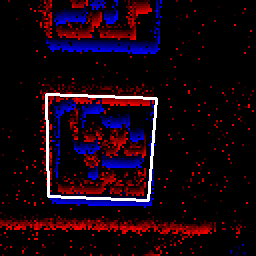}
\includegraphics[width=0.17\textwidth]{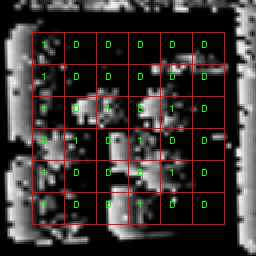}
\includegraphics[width=0.17\textwidth]{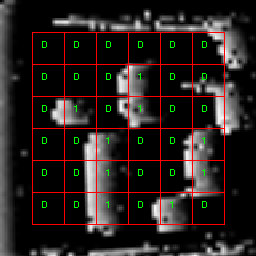}
\includegraphics[width=0.17\textwidth,frame]{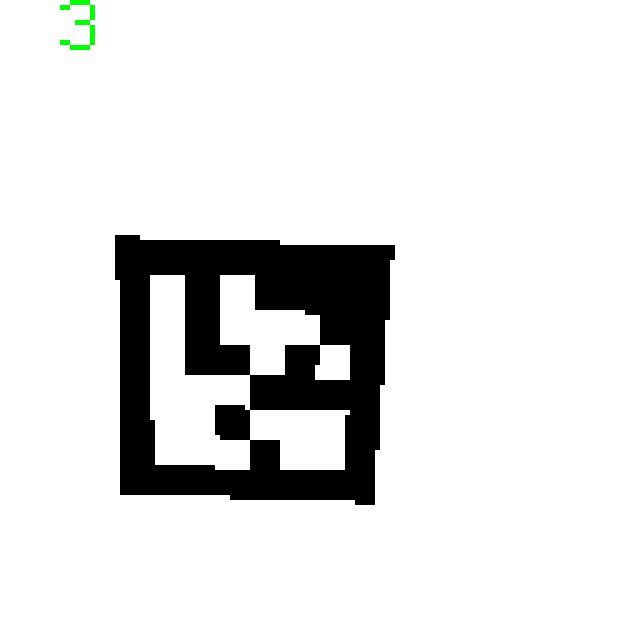}
\\
\raisebox{1.5cm}{6  } \subcaptionbox{\label{subfig:rgb}}{\raisebox{.5cm}{\includegraphics[width=0.19\textwidth]{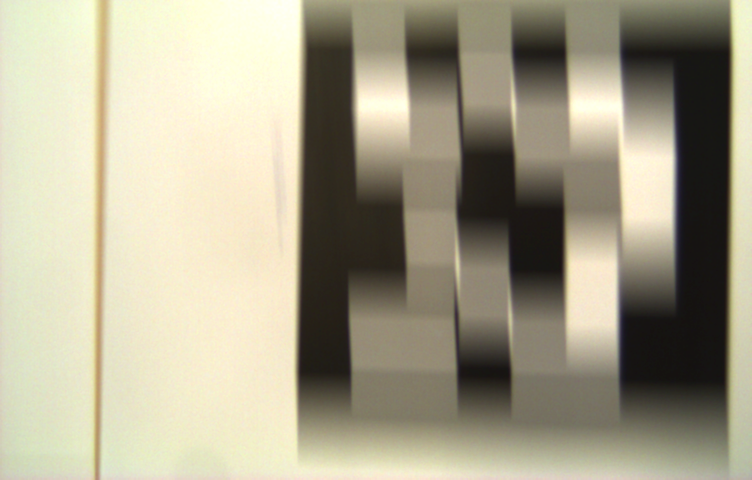}}}
\subcaptionbox{\label{subfig:candidate}}{\includegraphics[width=0.17\textwidth]{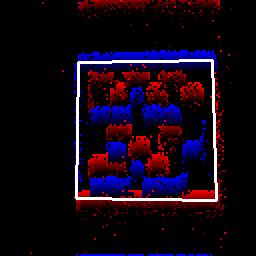}}
\subcaptionbox{\label{subfig:on_decoding}}{\includegraphics[width=0.17\textwidth]{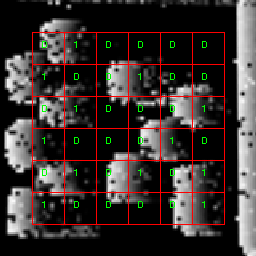}}
\subcaptionbox{\label{subfig:off_decoding}}{\includegraphics[width=0.17\textwidth]{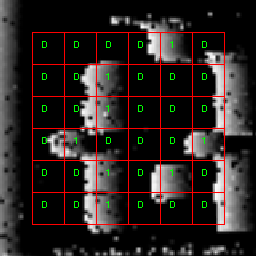}}
\subcaptionbox{\label{subfig:reconst}}{\includegraphics[width=0.17\textwidth,frame]{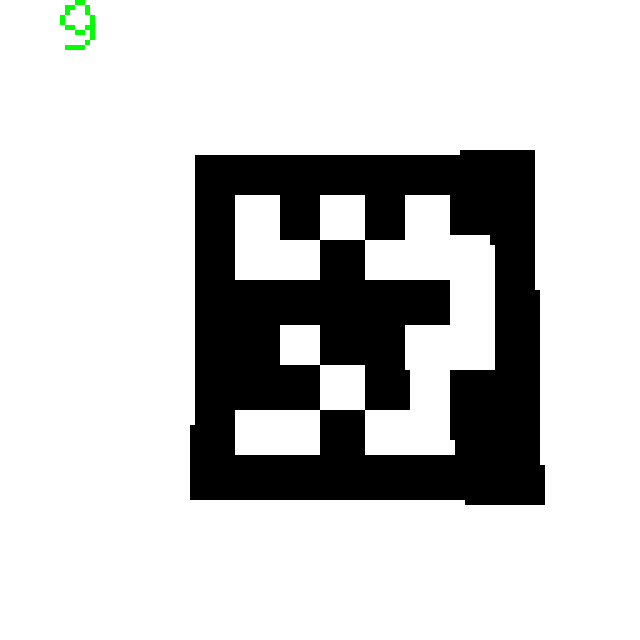}}

    \caption{Examples of successful detections in our dataset. For each case, the image from the RGB camera (\subref{subfig:rgb}), the candidate overlaid on the events (\subref{subfig:candidate}), Gaussian responses overlaid on unwarped \emph{on} (\subref{subfig:on_decoding}) and \emph{off} (\subref{subfig:off_decoding}) events, and finally the reconstruction of the marker in event camera's field of view (\subref{subfig:reconst}) is presented. Please note that \emph{on} and \emph{off} events in column (\subref{subfig:candidate}) are shown in red and blue colors respectively.}
    \label{fig:successes}
\end{figure*}

\subsection{Qualitative Evaluation and Discussion}
For qualitative evaluation, we visualize different steps of our marker detection algorithm for successful and unsuccessful cases. First, successful cases for different markers are presented in Figure \ref{fig:successes}. These cases are drawn from the sequences used for quantitative evaluation where the marker is detectable in the color camera as well as the event camera. The picture from the color camera, the marker candidate, normalized thresholded Gaussian responses for \emph{on} and \emph{off} events, and the marker reconstruction are visualized for each case in columns (a) to (e). The marker is moving horizontally with respect to the camera in rows 1 to 3 and vertically in rows 4 to 6. As can be seen, despite the high frame rate of the color camera ($60$ fps) the images are very blurry. As established in the quantitative results section, in no one of the color images of Figure \ref{fig:successes}, the  ArUco algorithm was able to detect the marker due to its high sensitivity to blurred images. The normalized thresholded Gaussian responses in columns (c) and (d) make it possible to see how the decoded marker is reconstructed.

\begin{figure*}[h!]
    \raisebox{1.5cm}{\parbox{2.5cm}{\centering all events}}\includegraphics[width=0.18\textwidth]{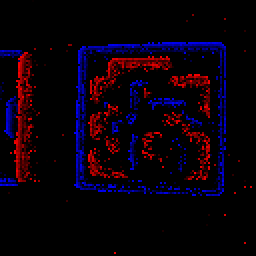}
    \includegraphics[width=0.18\textwidth]{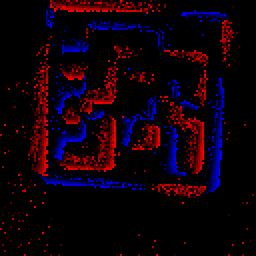}
    \includegraphics[width=0.18\textwidth]{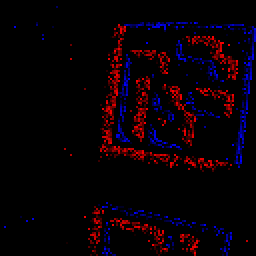}
    \includegraphics[width=0.18\textwidth]{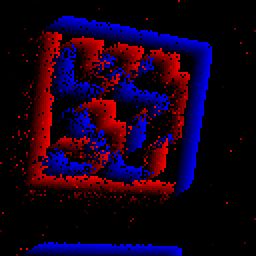}
    \\
    \raisebox{1.5cm}{\parbox{2.5cm}{\centering line segments\\ (\emph{on} events)}}\includegraphics[width=0.18\textwidth]{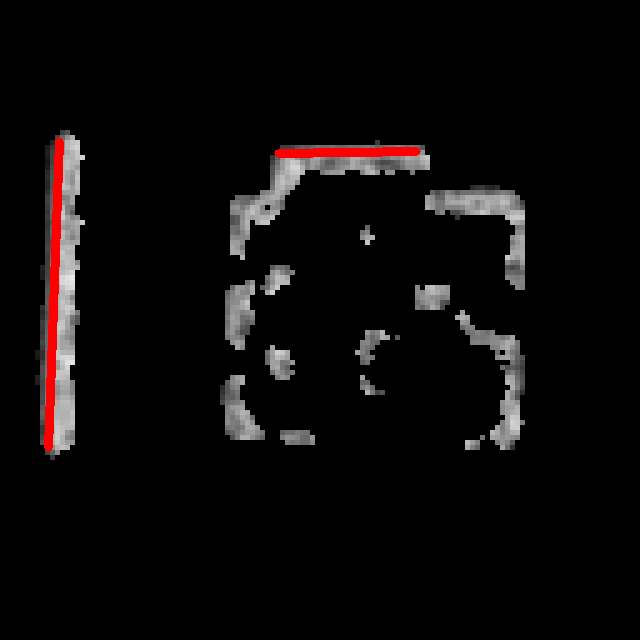}
    \includegraphics[width=0.18\textwidth]{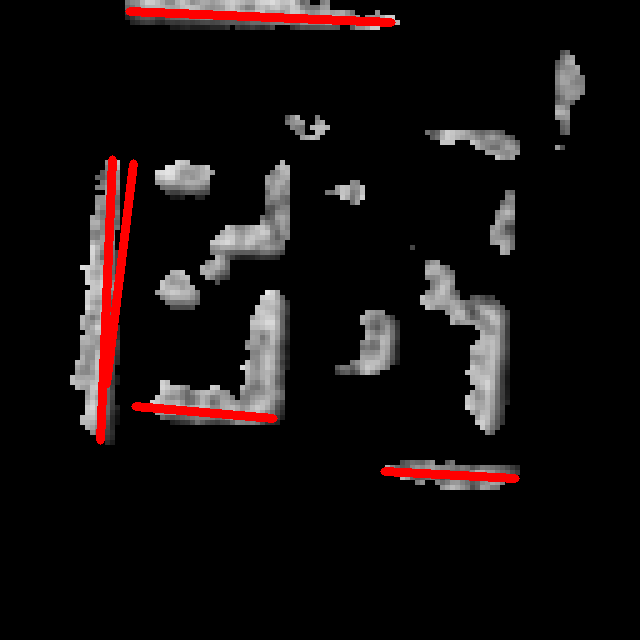}
    \includegraphics[width=0.18\textwidth]{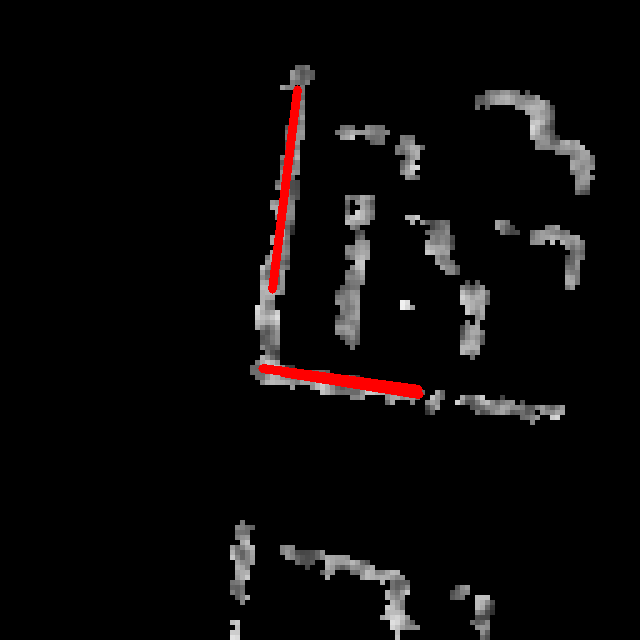}
    \includegraphics[width=0.18\textwidth]{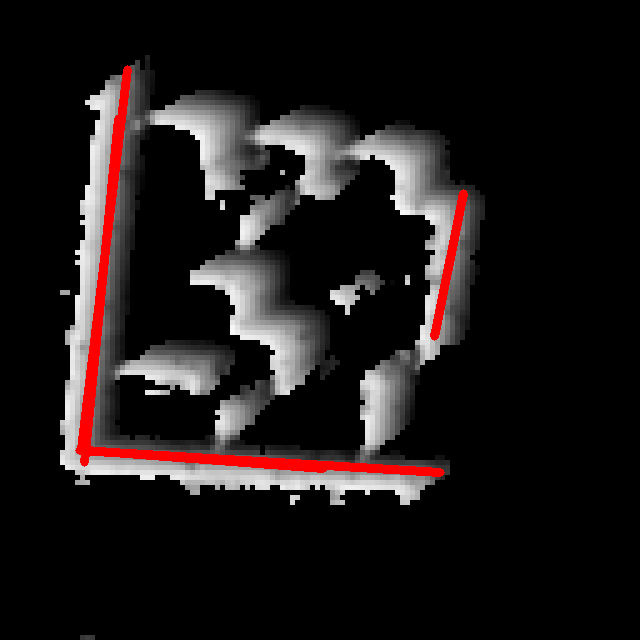}
    \\
    \parbox{2.5cm}{\centering line segments\\(\emph{off} events)}\parbox{0.18\textwidth}{ \includegraphics[width=0.18\textwidth]{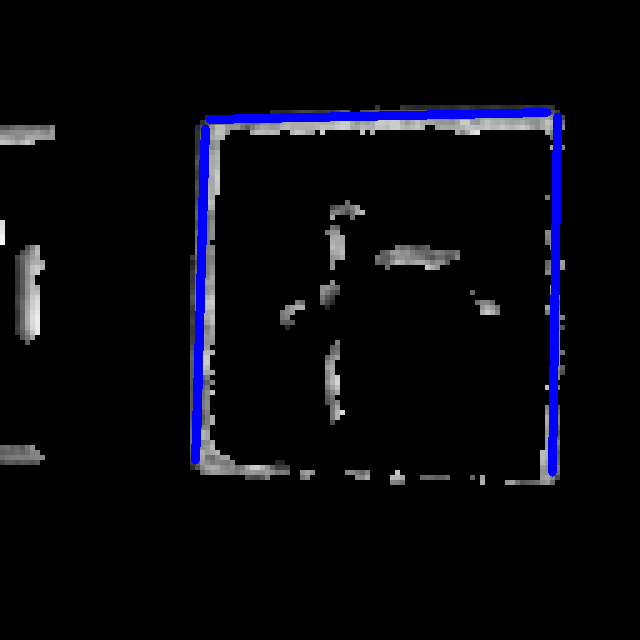}\subcaption{zooming}\label{subfig:zooming}}
    \parbox{0.18\textwidth}{ \includegraphics[width=0.18\textwidth]{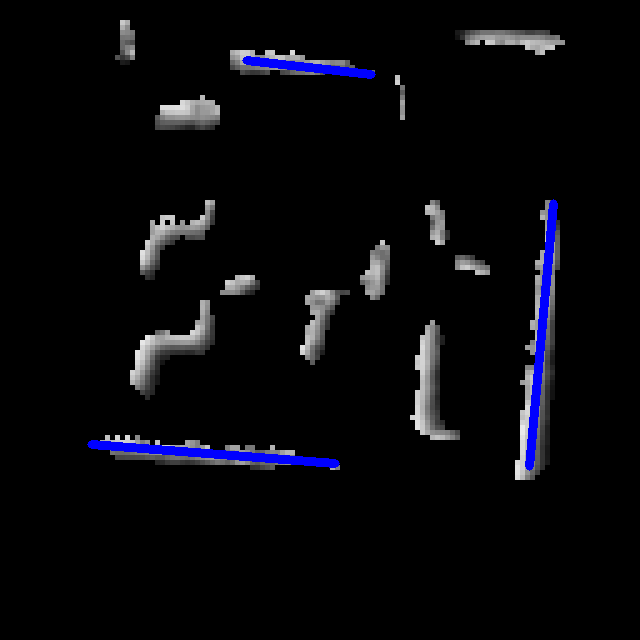}\subcaption{rotating}\label{subfig:rotating}}
    \parbox{0.18\textwidth}{ \includegraphics[width=0.18\textwidth]{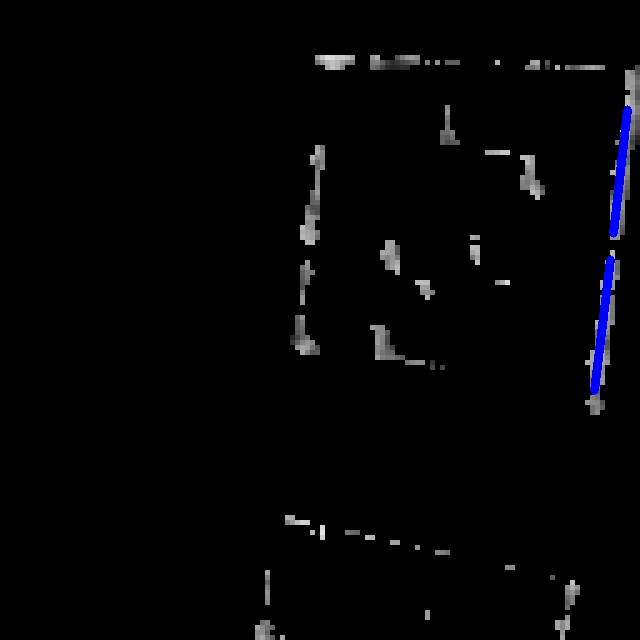}\subcaption{slow movement}\label{subfig:slow_movement}}
    \parbox{0.18\textwidth}{ \includegraphics[width=0.18\textwidth]{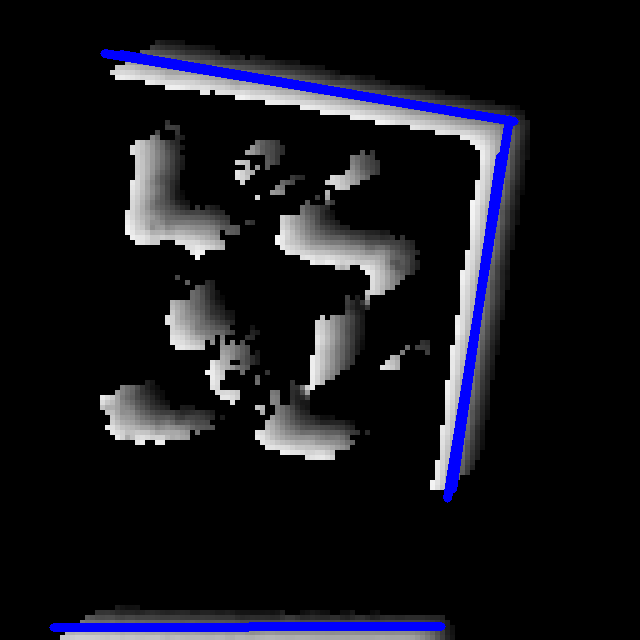}\subcaption{diagonal movement}\label{subfig:diagonal}}
    \caption{Cases where marker candidates cannot be formed properly because of unsuitable line segment detection. These include: (\subref{subfig:zooming}) zooming into or out of the marker, (\subref{subfig:rotating}) rotating with the center of rotation within the marker, and slow (\subref{subfig:slow_movement}) or diagonal (\subref{subfig:diagonal}) movement of the marker.}
    \label{fig:fail_movements}
\end{figure*}

\begin{figure*}[h!]
    \centering
    \raisebox{1.5cm}{1  }
    \includegraphics[width=0.17\textwidth]{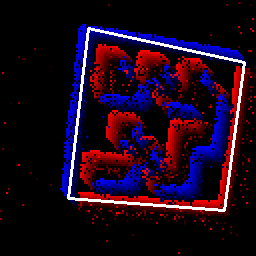}
    \includegraphics[width=0.17\textwidth]{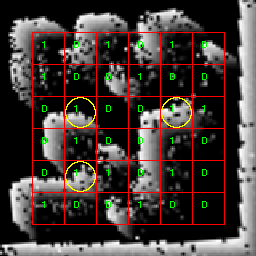}
    \includegraphics[width=0.17\textwidth]{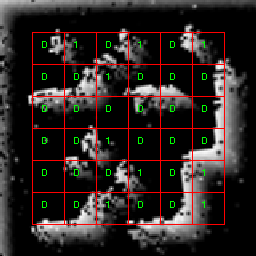}
    \includegraphics[width=0.17\textwidth]{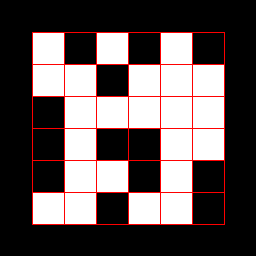}
    \includegraphics[width=0.17\textwidth]{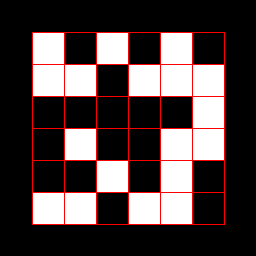}\\
    \raisebox{1.5cm}{2  }
    \subcaptionbox{}{\includegraphics[width=0.17\textwidth]{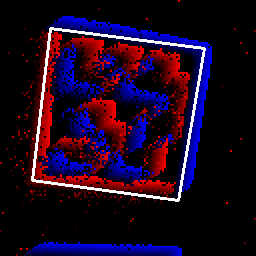}}
    \subcaptionbox{}{\includegraphics[width=0.17\textwidth]{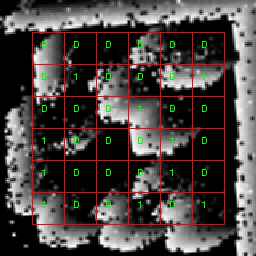}}
    \subcaptionbox{}{\includegraphics[width=0.17\textwidth]{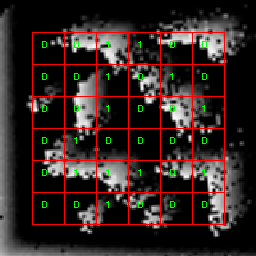}}
    \subcaptionbox{}{\includegraphics[width=0.17\textwidth]{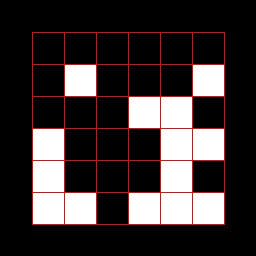}}
    \subcaptionbox{}{\includegraphics[width=0.17\textwidth]{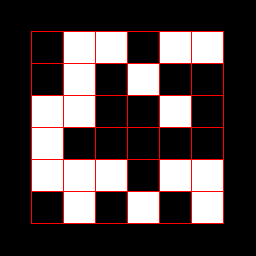}}
    \caption{Problems with diagonal movement in marker decoding. Problems with leaking events and candidate edges that are too long are shown in the first and second row respectively. In each case the marker candidate (a), thresholded normalized Gaussian responses for \emph{on} (b) and \emph{off} (c) events, the decoded marker (d), and the correct marker (e) are visualized.}
    \label{fig:diag_failures}
\end{figure*}

Four different types of movements for which our detection algorithm cannot function properly are shown in Figure \ref{fig:fail_movements}. In each category you can see the event image for \emph{on} (red) and \emph{off} (blue) events in the first row. The smoothed event image and line segment detections are shown in the second and third row separately for \emph{on} and \emph{off} events. In the first column (a) the camera is zooming out from the marker. As can be observed, there are no line segment detections in the \emph{on} events representing any of the marker edges. This is because when the zoom center is within the marker all its edges appear to move away/towards the zoom center. Hence marker edges would not be detectable for either \emph{on} or \emph{off} events and no candidate can be formed. The second column (b) shows a marker rotating with the center of rotation within the marker. As can be seen, each edge of the marker produces both \emph{on} and \emph{off} events in different places. This prevents the whole edge to be detected in either type of event which prevents the formation of a correct candidate. Column (c) represents the case where the camera is moving too slowly. This results in the production of few events which in turn makes it impossible to detect marker edges using line segments properly. Finally, in the last column (d) the line segment detections for a marker moving diagonally are shown. Although it might seem fine in the pictures, the detected line segments turn out to be too long for \emph{off} events. This is because the \emph{off} events from two neighboring edges combine which makes these edges appear longer. This becomes problematic when the marker is unwarped for decoding. This is shown in the second row of Figure \ref{fig:diag_failures}. The distortion in unwarping the events image in column (b) row 2 is especially noticeable if attention is paid to the \emph{on} events formed on the top edge. This distortion makes the events move to the wrong cell and finally being wrongly decoded as is visible in the decoded result in column (d) of the second row.
There is another problem with diagonal movement that could happen even if the marker candidate is formed properly. This is visualized in the first row of Figure \ref{fig:diag_failures}. The issue is that when moving diagonally events from cells below can leak into cells above or vice versa. This could change the result of Gaussian convolutions and change the decoded color of a cell. More specifically the events leaked from the lower left or upper left cells can make it appear that there is a change of color from the left cell to the current cell and hence produce a wrong color for the current cell in the decoding process. The affected cells for our example are indicated with yellow circles in column (b) and the first row of Figure \ref{fig:diag_failures}. As you can see in column (d) of the first row the resulted decoded marker has changed significantly.

As mentioned before, it is possible to use a intensity image reconstruction method to convert the events into intensity image in real-time and then apply a traditional marker detection algorithm. However, the good quality methods need a powerful dedicated GPU. As the first attempt to detect and decode binary planar markers only using the CPU unit in real-time, our method has the limitations mentioned in this section. Nevertheless, we think that it is possible to alleviate these problems by taking into account extra parameters such as the motion vector of the marker. However this is left to be done in future works.

\section{Conclusion}
\label{sec:conclusion}

A method for decoding square binary markers from the output of an event camera has been proposed in this paper. The algorithm can run in real-time on a single CPU core without the need for specialized hardware such as dedicated GPUs. To the best of our knowledge,  this is the first attempt to decode fiducial planar markers using event cameras. An additional contribution of this paper is that the different steps we have proposed for processing events can be helpful in the design of solutions for other problems using these cameras.

Experimental results show that our method is much superior compared to the RGB marker detector in the case of ArUco for fast-moving markers. This is because the intensity-based methods assume low blur in the image while we demonstrated that even with a high frame rate there can be a significant motion blur. Hence, the proposed method is ideal in settings where objects move very fast in directions roughly parallel or perpendicular to the camera, such as a production line.

On the other hand, for situations other than fast lateral (or up/down) movements our algorithm could be improved. As future work, we propose to work on these cases including diagonal movement, rotations, and zooming of the camera. We suggest it might be possible to overcome these limitation taking into account extra parameters such as the motion vector. Another matter that could be tested in future work is how robust is the method in different lighting conditions and in the presence of ambient light noise.

\section*{Acknowledgments} 

This project has been funded under projects TIN2019-75279-P and IFI16/00033 (ISCIII) of Spain Ministry of Economy, Industry and Competitiveness, and FEDER.

In addition, this work was done in collaboration with the Automation \& Robotics research group at the Interdisciplinary Centre for Security, Reliability and Trust (SnT), University of Luxembourg.

\bibliographystyle{ieeetr}
\bibliography{references}

\end{document}